\pdfoutput=1
\documentclass[11pt]{article}

\usepackage[final]{acl}

\usepackage{amsmath}
\usepackage{adjustbox}
\usepackage{booktabs}
\usepackage{comment}
\usepackage{latexsym}
\usepackage{multirow}
\usepackage{paralist}
\usepackage{placeins}
\usepackage{tabularx}
\usepackage{times}

\usepackage{float}       % Enables [H]
\usepackage{mdframed}
\usepackage{listings}
\usepackage[dvipsnames]{xcolor}
\usepackage{caption}

\usepackage{fancyvrb}
\usepackage{fvextra}

\definecolor{promptbg}{RGB}{250, 250, 252}
\definecolor{promptborder}{RGB}{210, 210, 230}
\definecolor{promptpurple}{RGB}{100, 50, 150}

\usepackage{xcolor}

\newif\ifhighlight
\highlightfalse   % comment this out or change to \highlightfalse to disable

\usepackage[T1]{fontenc}
\usepackage[utf8]{inputenc}

\usepackage{microtype}

\usepackage{inconsolata}

\usepackage{graphicx}

\title{Distilling Examples into Task Instructions: Enhanced In-Context Learning for Real-World B2B Conversations}

\author{Guy Rotman, Adi Kopilov, Danit Berger Zalmanson, Omri Allouche \\
  Gong.io \\
  \texttt{\{guy.rotman, adi.kopilov, danit.berger, omri.allouche\}@gong.io}}

\begin{document}
\maketitle

\begin{abstract}
    In-context learning (ICL) is the standard method for low-resource classification, yet its efficacy in specialized domains remains largely unexplored. We address the challenge of classifying semantically complex, multi-party B2B conversations, where traditional ICL encounters significant limitations, especially as context length increases due to the concatenation of multiple few-shot examples. We introduce the \texttt{Call Playbook} dataset, featuring five classification tasks derived from real-world B2B conversations targeting core sales concepts. To bridge the gap between performance and practical utility, we propose novel knowledge extraction methods that distill verbose examples into compact, interpretable representations of structured classification criteria and precise task descriptions. Our approach achieves a 99\% reduction in token usage and improves macro-averaged AUC by up to 7\% over traditional ICL. Notably, it remains robust as context grows, unlike advanced token compression baselines which degrade by over 9 F1 points. Importantly, our framework enables direct refinement of classification logic, addressing critical needs for transparency, efficiency, and user interaction in real-world NLP applications.
    \footnote{Data and code are available at: \url{https://github.com/gong-io/call-playbook}.} \footnote{Accepted for publication in Findings of the Association for Computational Linguistics 2026.}
\end{abstract}

\section{Introduction}
Recent advances in large language models (LLMs) have transformed natural language processing (NLP), particularly through in-context learning (ICL), where models perform tasks by conditioning on a few examples without parameter updates \cite{brown_language_models_2020,min_rethinking_2022}. This paradigm, commonly implemented through few-shot learning approaches, has demonstrated remarkable success across diverse NLP tasks \cite{sanh_hal_2022, chowdhery2023palm}. However, its application to specialized domains such as business-to-business (B2B) sales communications presents unique challenges that have yet to be thoroughly examined \cite{gupta2022leveraging, chamieh-etal-2024-llms}.

In B2B sales environments, analyzing prospect conversations is essential to extract actionable intelligence guiding deal strategy \cite{grosz1995discourse,dean_science_2017}. To achieve this at scale, organizations must automatically classify conversational segments across diverse and evolving task intents that are not known in advance. These tasks operate under severe constraints: limited labeled data, minimal annotation overhead, and the impracticality of fine-tuning models for each emerging intent.

Despite its theoretical suitability for these constraints, ICL faces significant practical hurdles. Current LLMs can potentially support context windows of up to millions of tokens \cite{reid2024gemini}; however, our analysis reveals severe performance degradation under standard ICL with far fewer tokens. This decline stems primarily from the concatenation of multiple few-shot examples, challenging the conventional assumption that increasing example count improves performance \cite{agarwal_many-shot_2024, bertsch_-context_2025}. These limitations are especially pronounced in the B2B domain, where professionals require interpretable and transparent outputs to audit decisions and maintain trust \cite{doshi-velez_towards_2017, lipton2018mythos}.

To address these challenges, we propose a knowledge extraction framework in which an LLM automatically derives generalizable task knowledge from a small set of labeled examples and converts it into coherent natural-language classification criteria and task descriptions, designed to substitute the few-shot examples in the prompt.

This approach is fundamentally distinct from existing strategies for improving ICL. Retrieval-based methods \cite{rubin_learning_2022} focus on selecting better examples to include in context, prompt optimization techniques restructure how examples are presented \cite{pan-etal-2023-context, ye_compositional_nodate}, and token-level compression methods reduce redundancy within individual examples, often yielding incoherent, fragmented tokens \cite{jiang-etal-2023-llmlingua,li-etal-2023-compressing}. In contrast, our approach eliminates the need for example concatenation by shifting to a different representational form. The resulting outputs are transparent and interpretable, supporting both automated processing and human oversight, making the approach uniquely suited for the efficiency, intent flexibility, and interpretability required in practical B2B applications.

To provide a basis for evaluation, we introduce five novel B2B classification tasks derived from real-world sales conversations. These tasks target prospect understanding across fundamental sales concepts and serve to highlight the limitations of standard ICL in high-stakes business contexts.

Our main contributions are as follows: (1) We introduce a new annotated B2B dataset spanning five core sales concepts. (2) We propose novel ICL methods that shift from example concatenation to automated knowledge extraction. (3) Extensive experiments demonstrate superior performance over standard few-shot baselines and advanced token compression methods. (4) Our approach yields significant computational savings through token reduction and faster inference times at scale. (5) The framework offers built-in interpretability, enabling human-in-the-loop enhancement and user guidance. (6) We support cross-model knowledge transfer, allowing large models to generate task instructions for deployment on smaller ones.

\begin{figure}[!ht]
  \includegraphics[width=\columnwidth]{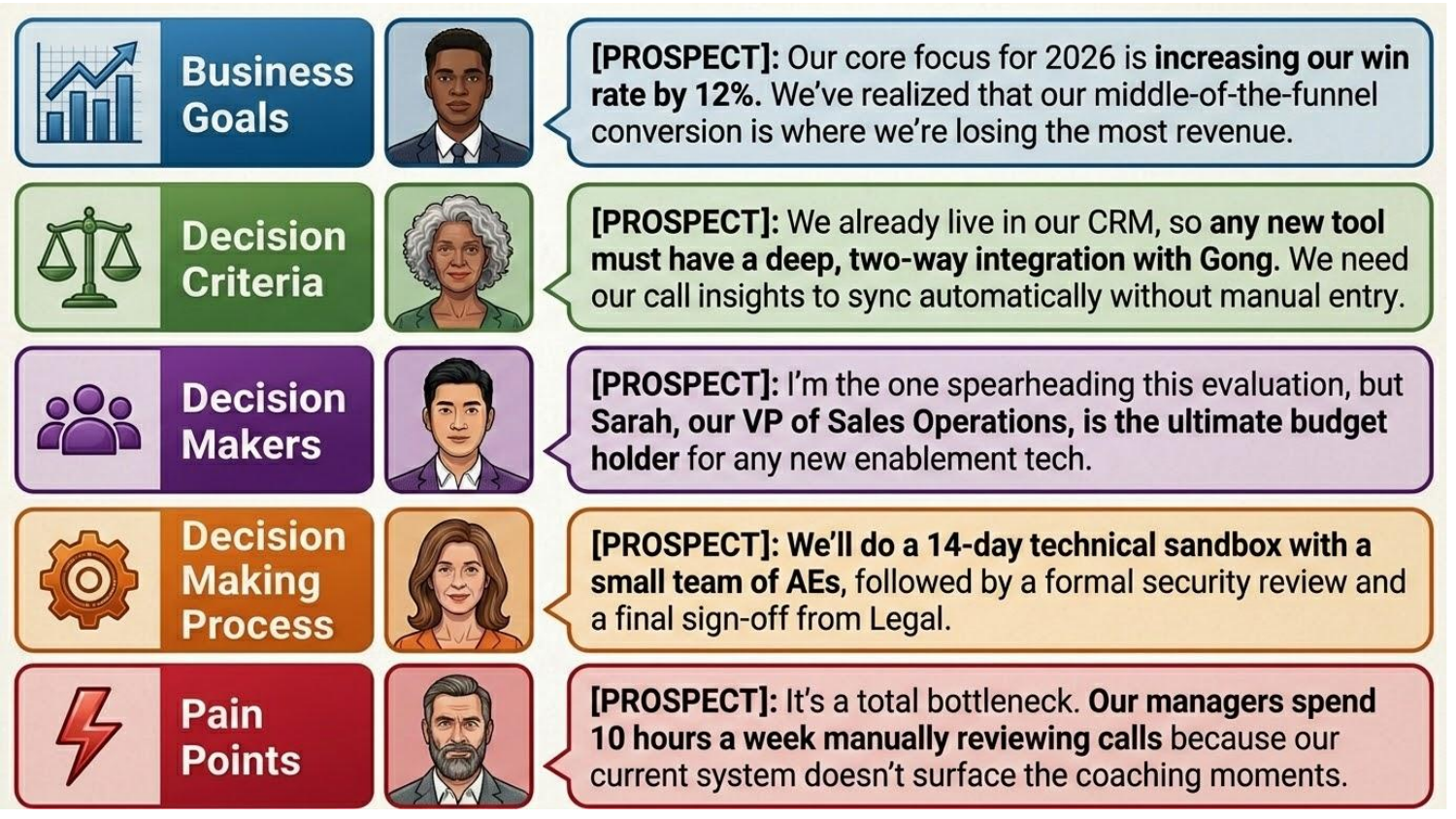}
\caption{\textbf{The Call Playbook Dataset.} Representative examples of prospect monologues mapped to the five targeted B2B sales concepts.}  \label{fig:call_playbook_avatar}
\end{figure}

\begin{table*}[!ht]
\centering
\begin{adjustbox}{width=\linewidth}
\begin{tabular}{|l|cccccc|cccccc|}
\hline
& \multicolumn{6}{c|}{\textbf{Train}} & \multicolumn{6}{c|}{\textbf{Test}} \\ \hline
\textbf{Dataset} & \textbf{Pos} & \textbf{Neg} & \textbf{Total} & \textbf{Calls} & \textbf{Avg Words} & \textbf{Prospect \%} & \textbf{Pos} & \textbf{Neg} & \textbf{Total} & \textbf{Calls} & \textbf{Avg Words} & \textbf{Prospect \%} \\ \hline
Business Goals & 100 & 100 & 200 & 25 & 288 & 48 & 100 & 100 & 200 & 25 & 279 & 51 \\ \hline
Decision Criteria & 86 & 114 & 200 & 25 & 277 & 50 & 94 & 106 & 200 & 25 & 274 & 47 \\ \hline
Decision Makers & 32 & 168 & 200 & 25 & 259 & 48 & 35 & 165 & 200 & 25 & 274 & 43 \\ \hline
Decision Making Process & 61 & 139 & 200 & 20 & 271 & 48 & 68 & 132 & 200 & 21 & 262 & 42 \\ \hline
Pain Points & 100 & 100 & 200 & 25 & 275 & 45 & 100 & 100 & 200 & 25 & 296 & 49 \\ \hline
\end{tabular}
\end{adjustbox}
\caption{\texttt{Call Playbook} statistics. Pos: positive examples; Neg: negative examples; Total: sample count; Calls: number of unique calls; Avg Words: average words per sample; Prospect \%: percentage of prospect-side dialogue per sample.}
\label{tab:dataset_statistics}
\end{table*}

\section{Related Work}

\subsection{In-Context Learning (ICL)}
In-context learning (ICL) has emerged as a major development in NLP, showcasing the ability of LLMs to perform diverse tasks using a few labeled samples provided in the prompt, without requiring explicit fine-tuning \cite{brown_language_models_2020, dong_survey_2024}. This capability, first demonstrated by models such as GPT-3, has spurred extensive research into its mechanisms, performance, and limitations.

Much of this research explores the role of demonstrations in ICL, focusing on prompt design and the selection and ordering of examples \cite{liu-etal-2022-makes, pan-etal-2023-context}. Strategies include retrieving semantically similar instances \cite{rubin_learning_2022} or modeling inter-example relationships \cite{ye_compositional_nodate}, as well as employing active learning to select effective subsets \cite{zhang_active_2022}. In contrast, this work derives task-specific knowledge directly from a small number of examples.

Recent studies have shown that ICL struggles with long contexts, especially when many examples are used \cite{li_long-context_2024, lee-etal-2025-ethic, modarressinolima2025}. To address this challenge, compression techniques have been explored for reducing example length in NLP applications. Token-level methods remove redundant content based on self-information metrics \cite{li-etal-2023-compressing} or perform iterative compression through instruction-tuned models \cite{jiang-etal-2023-llmlingua,pan-etal-2024-llmlingua}, but are limited to selecting subsets of existing tokens, often resulting in incoherent examples.

Model-based approaches train dedicated compression models for specific contexts, such as abstractive summarization for retrieval-augmented generation \cite{xu2023recomp} or active compression for question answering \cite{yoon-etal-2024-compact}. While effective, these methods require labeled datasets and specialized model training. Unlike these approaches, our method generates concise, interpretable demonstrations that preserve coherence while distilling essential task knowledge without requiring additional model training.

\subsection{NLP for the B2B Domain}
B2B conversations pose distinct challenges for NLP due to their complexity, involving multiple stakeholders, long sales cycles, and domain-specific language \cite{grewal2022business, wu2024intelligentcustomer}. Business signals are often implicit \cite{voria2024recover}, positioning B2B conversations as a rigorous benchmark for evaluating how LLMs interpret nuanced, context-specific language.

NLP is increasingly applied in the B2B domain for business goal identification \cite{campbell_expertise_nodate,spruit_automated_2021}, sales enhancement \cite{patel2022enhancing}, understanding sales conversations \cite{chai2001role}, customer segmentation \cite{lieder_learning_2019}, and sales forecasting \cite{bohanec_organizational_2017, zahid_predicting_2021, rohaan_using_2022}. However, existing work lacks consideration of the fluid multi-stakeholder dynamics and the evolving context throughout deal progression.

To bridge this gap, this paper presents a novel dataset of authentic B2B sales conversations from real-world interactions with multiple stakeholders across different deal stages, annotated for diverse interconnected prospect-focused business concepts. Unlike existing B2B datasets that typically focus on customer support tickets, product reviews, or short transactional exchanges, our dataset captures extended multi-party sales dialogues that reflect the complexity of actual business negotiations.

In this work, we demonstrate how LLMs can automatically generate user-guiding task instructions to enhance ICL, enabling comprehensive analysis of prospect communication patterns throughout these extended dialogues.

% Prior work spans broad marketing analyses \cite{han_artificial_2021} to specific tasks such as contract classification \cite{braun_clause_2022}. We extend this line by targeting the extraction of fine-grained business signals from real B2B conversations.

%\footnote{\textcolor{Blue}{It's not clear from this section what exactly makes the B2B domain complex/interesting. Is it the domain itself and its nuances? Or perhaps the fact that these are sales conversations, making the dataset challenging due to its semantic variability? I would expand on this in a paragraph to clarify why this topic is particularly interesting, and how it highlights a more general issue — semantic variability within domain-specific knowledge that an LLM is expected to handle, in my view}}

\begin{figure*}[t]
  \includegraphics[width=\textwidth]{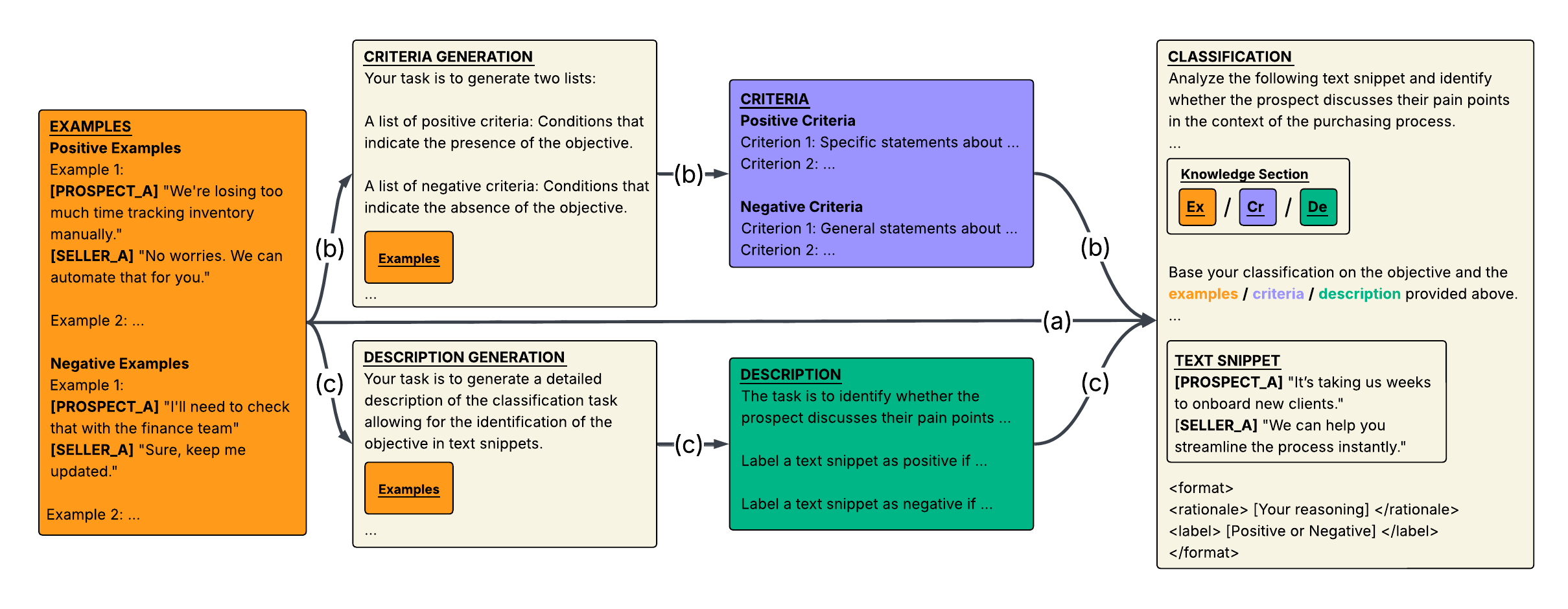}
  \caption{Classification framework overview. The process begins by sampling labeled examples (a), which are either used directly in the knowledge section (traditional few-shot learning) or transformed through knowledge extraction into criteria lists (b) or a task description (c). This extracted knowledge replaces raw examples in the classification prompt, producing structured predictions with explicit rationales for enhanced classification performance.}
  \label{fig:main_figure}
\end{figure*}

\section{The Call Playbook Dataset}

This section describes the data collection, text processing, and annotation processes used to construct the \texttt{Call Playbook} dataset.

\subsection{Dataset Overview}

We constructed the \texttt{Call Playbook} dataset from 50~English B2B sales calls with durations ranging from 30~to 90~minutes. Each transcript is structured as a sequence of monologues attributed to speakers on either the seller or the prospect side. The dataset contains annotations for five key sales concepts targeting core prospect dimensions. Each concept captures a distinct aspect of the prospect's intent and purchasing process:
\begin{compactitem}
    \item \textbf{Business Goals} describe the outcomes or objectives the prospect wishes to achieve.
    \item \textbf{Decision Criteria} refer to the standards used to evaluate potential solutions.
    \item \textbf{Decision Makers} identify individuals or roles involved in making the purchasing decision.
    \item \textbf{Decision Making Process} refers to the sequence of steps the prospect follows when making a decision.
    \item \textbf{Pain Points} reflect the challenges and obstacles the prospect seeks to address.
\end{compactitem}
Figure~\ref{fig:call_playbook_avatar} provides a conceptual illustration of these categories and representative examples of annotated prospect monologues, with relevant concept text highlighted in bold.

\subsection{Data Annotation}

The annotation task involved three trained in-house annotators who identified and labeled textual spans where each concept was expressed, guided by a domain expert who supervised the process. Each call was segmented into overlapping \textit{snippets} of five consecutive monologues with an overlap of one monologue between adjacent snippets.\footnote{Short monologues containing fewer than five words are excluded from the count.} This segmentation produces dense, contextually grounded segments that preserve the flow of conversation while enabling targeted classification. Each snippet was then assigned binary labels for the five concepts: a snippet was marked \textit{positive} if it contained at least one annotated span for a concept, and \textit{negative} otherwise. Any disagreements that occurred between annotators were resolved through discussion until consensus was reached.

For each concept, we created class-balanced train and test sets containing 200~samples each. When positive examples were limited, we evenly split them between sets and filled the remainder with randomly sampled negatives. To avoid data leakage, we ensured calls did not overlap between sets. This balanced construction establishes a rigorous benchmark that prevents evaluation metrics from being masked by majority-class prevalence.

Table~\ref{tab:dataset_statistics} summarizes key statistics for the \texttt{Call Playbook} dataset, including the number of positive and negative examples, total sample count, number of unique calls, average number of words per sample, and the proportion of dialogues attributed to the prospect. The dataset maintains a balanced or nearly balanced distribution across most categories and contains detailed conversational examples of business interactions.

Data curation procedures and usage guidelines are outlined in Appendix~\ref{appsec:data_curation_and_guidelines}. To protect sensitive information, the dataset was thoroughly processed and anonymized as detailed in Appendix~\ref{appsec:data_anonymization}. Authentic examples illustrating the diversity and complexity of the dataset are provided in Appendix~\ref{appsec:dataset_examples}.

\section{Methodology}
\label{sec:methodology}

This section presents our approach to enhancing the effectiveness of ICL for classification tasks in B2B conversational analysis. We begin by formalizing the problem setup and then describe our novel methods for improving few-shot classification performance within this framework.

\subsection{Problem Setup}
\label{subsec:problem_setup}
To reflect practical constraints in conversational analysis, we consider the problem of classifying conversational segments into domain-relevant categories under conditions of minimal task specifications and few labeled examples. 

We define this setup formally as follows: Given a short user-provided task intent $i \in I$ and a labeled dataset $E = \{(x_j, y_j)\}_{j=1}^M$, where each $x_j \in X$ is a conversational segment and $y_j \in C$ is its corresponding class label, our goal is to construct from $E$ a classifier $f: I \times X \rightarrow C$ that accurately predicts the label for an unseen segment $x_{\text{test}}$ given an intent $i$. In our experimental setup, we focus on binary classification where $C = \{0, 1\}$, with $1$ indicating the presence of the target concept as specified by $i$, and $0$ indicating its absence.

This classification task presents multiple challenges that make ICL particularly suitable for practical B2B applications: (1)~\textbf{Limited data availability}: specialized domains often have scarce labeled examples due to their specific nature and privacy constraints; (2)~\textbf{Minimal user overhead}: practitioners need quick deployment without extensive annotation or model training; (3) \textbf{Task diversity}: user intents vary widely, making task-specific fine-tuning of multiple models impractical at scale.

ICL addresses these constraints by leveraging a general
LLM to classify diverse task intents using only a few examples. Our framework extends this capability by extracting interpretable knowledge that enables human-in-the-loop intervention.

\subsection{Classification Process}
\label{subsec:classification_process}

Our general classification framework, illustrated in Figure~\ref{fig:main_figure}, proceeds as follows:

\noindent\textbf{Step 1: Sample Selection.} We randomly sample a small subset of $N$ labeled examples from the training set $E$, preserving the original class distribution, to serve as the basis for ICL. This step is shared across all proposed methods.

\noindent\textbf{Step 2: Knowledge Extraction.} Based on these examples, we either use them directly in a standard few-shot prompt or transform them into distilled knowledge, such as criteria or detailed descriptions, using an LLM. When employed, we use the same LLM for this step and for the classification task described in Step 4 for consistency.

\noindent\textbf{Step 3: Prompt Construction.} We construct a classification prompt that includes the user intent $i$, either the sampled examples or the derived knowledge, and the test snippet $x_{\text{test}}$ to be classified. For all methods, we use the same classification prompt template, further discussed in Appendix~\ref{appsec:classification_methodology}.

\noindent\textbf{Step 4: Classification.} We query an LLM with the constructed prompt to obtain a classification prediction $\hat{y} \in C$ for each test snippet.

Rather than relying on raw examples in the prompt, our method introduces a knowledge extraction phase that summarizes them into generalizable knowledge. This not only overcomes token limitations in few-shot setups but also supports strong performance and improved user interpretability.

\subsection{Classification Methods}
\label{subsec:methods}

\subsubsection{Few-shot Learning (Examples)}
\label{subsubsec:few_shot}

The standard few-shot learning approach, which we denote as ``Examples'', serves as our main baseline. In this method, we directly include the $N$ sampled examples in the classification prompt to provide guidance for the model, as shown in Figure~\ref{fig:examples_knowledge_section}.

While this approach has proven effective for many tasks, it faces notable challenges in conversational domains. Each snippet can be hundreds of words long (see Table~\ref{tab:dataset_statistics}), quickly consuming token budget as $N$ increases. Also, the model must infer classification rules implicitly from examples, which may lead to poor generalization, especially when the task is difficult or underspecified.

\subsubsection{Summary-Ex Method}
\label{subsubsec:summary_ex}

To harness the advantages of information compression and reduced contextual noise, we introduce ``Summary-Ex'' (Summary from \textbf{Ex}amples) as an intermediate baseline. This method replaces the full examples with concise summaries that retain essential discriminative information:\\
\noindent\textbf{Step 2: Example Summarization.} We prompt an LLM to generate a brief summary for each example, condensing it to 3-5 sentences while preserving the original conversation format and speaker affiliations. The summarization process focuses on removing redundant information and filler words while retaining all discussed business content and maintaining the essential structure and flow of the conversation. The detailed prompt for summarization is provided in Appendix~\ref{appsubsec:summary_generation}.

\noindent\textbf{Steps 3-4: Classification with Summarized Examples.} We substitute the original sampled examples with their summarized variants within the standard few-shot prompt format to classify the original text snippets.

This approach reduces token usage compared to full examples while maintaining the intuitive example-based learning paradigm. However, it still requires the model to implicitly infer classification patterns and may lose important contextual nuances during the summarization process.

\subsubsection{Criteria-Ex Method}
\label{subsubsec:criteria_ex}

Our first alternative, ``Criteria-Ex'' (Criteria from \textbf{Ex}amples), extracts explicit classification criteria from training examples rather than relying on implicit pattern inference:

\noindent\textbf{Step 2: Knowledge Extraction.} We instruct the relevant LLM to generate two lists of criteria based on the sampled examples and the user intent $i$: (1) a list of positive criteria indicating the presence of the concept in the text snippet, and (2) a list of negative criteria indicating its absence. The prompt (detailed in Appendix~\ref{appsubsec:criteria_generation}) directs the model to analyze the distinguishing patterns between positive and negative examples. For binary B2B tasks, this entails identifying patterns that indicate whether a concept is discussed or not in the conversation.

\noindent\textbf{Steps 3-4: Classification with Criteria.} We proceed with classification by replacing the few-shot examples in the classification prompt with the generated criteria (see Figure~\ref{fig:criteria_knowledge_section}).

This approach significantly reduces token usage compared to few-shot examples, as the criteria typically require far fewer tokens than the original conversation snippets. It also enhances explainability by making classification logic explicit rather than implicit, and improves generalization by extracting patterns rather than relying on specific examples.

\subsubsection{Description-Ex Method}
\label{subsubsec:description_ex}

Our next method, ``Description-Ex'', similarly transforms \textbf{Ex}amples into a detailed task description that extends beyond the short user intent:

\noindent\textbf{Step 2: Knowledge Extraction.} We prompt the relevant LLM to analyze the sampled examples and user intent $i$ to generate a comprehensive task description that captures the essence of the classification task. The description explains the characteristics that indicate the presence or absence of the target concept, and provides a coherent explanation of the concept boundaries. The prompt for generating descriptions is provided in Appendix~\ref{appsubsec:description_generation}.

\noindent\textbf{Steps 3-4: Classification with Description.} We replace the few-shot examples with the generated description in our classification prompt (see Figure~\ref{fig:description_knowledge_section}) and proceed with classification.

This approach offers benefits similar to Criteria-Ex but provides a cohesive narrative explanation that may better capture complex relationships and context-dependent aspects of classification. Its structured format aligns more naturally with how LLMs process instructions, which can improve generalization in challenging scenarios where rigid criteria may overlook important subtleties.

\subsubsection{Iterative Improvement Methods}
\label{subsubsec:iterative_methods}

We further investigate whether our proposed methods can be enhanced through iteration, examining the potential for refining derived knowledge.

\textbf{Criteria-De} generates classification criteria from a previously generated task  \textbf{De}scription, produced by Description-Ex, rather than directly from examples. The associated prompt is detailed in Appendix~\ref{appsubsec:criteria_generation}. 

\textbf{Description-Cr} generates a task description from previously generated  \textbf{Cr}iteria, produced by Criteria-Ex, testing whether structured criteria can be expanded into a more comprehensive narrative. The associated prompt is detailed in Appendix~\ref{appsubsec:description_generation}.

These iterative variants illustrate how knowledge representations can be progressively refined, e.g., from examples to descriptions to criteria and back, revealing their complementary roles. This capacity for stepwise enhancement renders the approach especially well-suited for dynamic, human-in-the-loop workflows, where evolving guidance or labeling needs are met through successive refinements rather than redesigning prompts from scratch.

\begin{figure*}[t]
  \includegraphics[width=\textwidth]{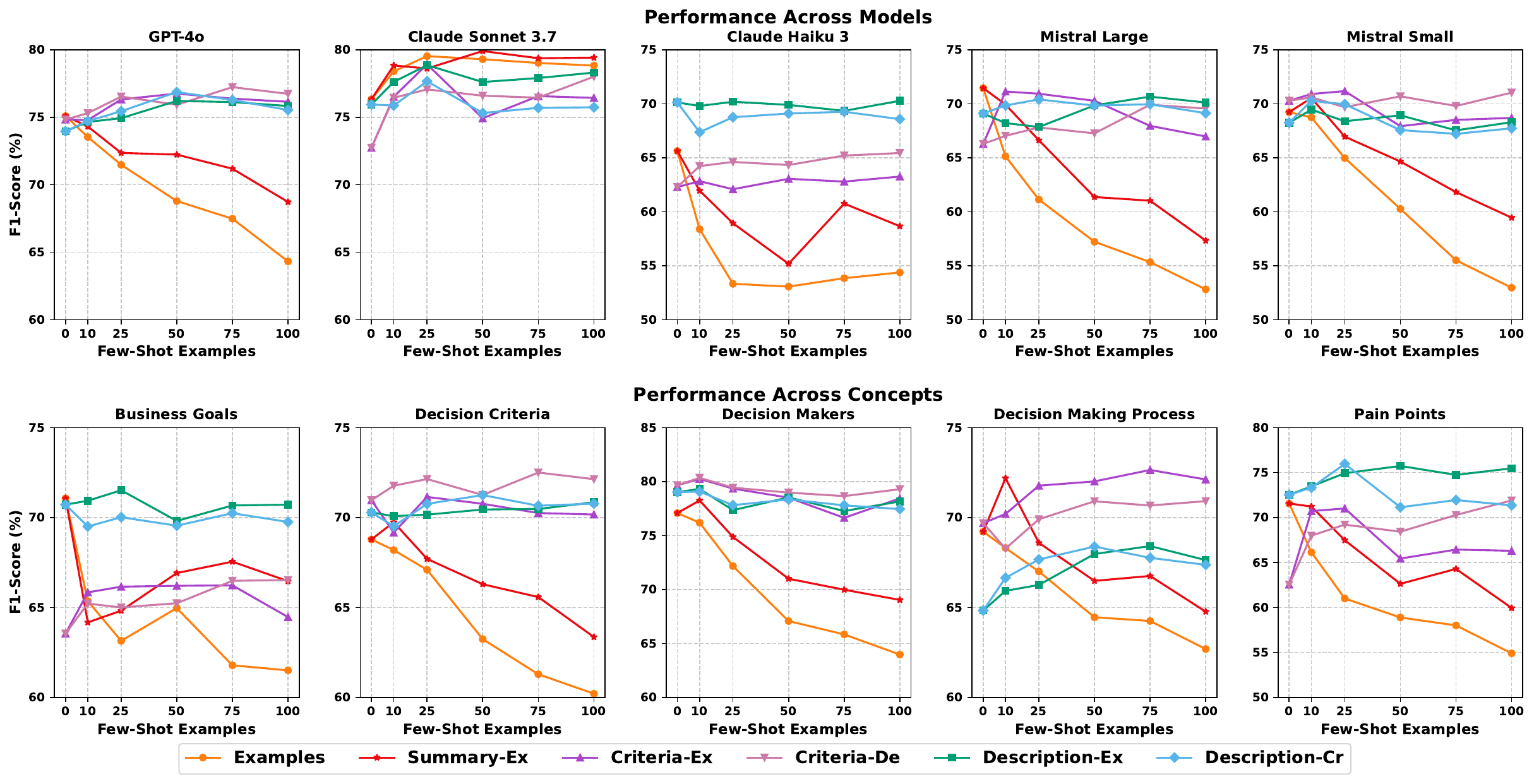}
  \caption{Top: Macro-average F1 performance for each model, averaged over all concepts. Bottom: Macro-average F1 performance for each concept, averaged over all models.}
  \label{fig:combined_performance}
\end{figure*}

\section{Experiments}
We evaluate all methods on the five concepts of \texttt{Call Playbook}. Using ICL, we systematically vary the number of examples ($0$, $10$, $25$, $50$, $75$, and $100$) to assess performance across different few-shot levels\footnote{In zero-shot scenarios, our criteria- and description-based methods distill information solely from the user intent, leveraging general knowledge to enable operation without examples.}.

Examples are randomly sampled while preserving the original class distribution. To mitigate potential biases and ensure statistical reliability, we repeat each configuration five times with different random samples and report averaged metrics.

Our evaluation covers five LLMs of varying capabilities: GPT-4o \cite{hurst2024gpt}, Claude Sonnet 3.7 and Claude Haiku 3 \cite{anthropic2024claude3}, as well as Mistral Large and Mistral Small \cite{mistral_paper}. This selection includes both proprietary and open-weight models with diverse architectures and parameter scales. To ensure a fair comparison, we use the same set of examples across all models and set \emph{temperature} to 0 for deterministic outputs.

For implementation, we use the \texttt{LangChain} framework\footnote{\url{https://www.langchain.com/}} with the Azure API for GPT-4o and the AWS Bedrock API for the remaining models.

\section{Results}

\subsection{Main Results}
\label{subsec:main_results}

Our ICL analysis across the five LLMs and five concepts reveals key classification performance patterns. Figure~\ref{fig:combined_performance} presents a comprehensive view of macro-average F1 scores, with the top row showing average performance across models and the bottom row displaying results across concepts. The complete results for all models and concepts are presented in Appendix~\ref{appsubsec:complete_grid}. Detailed qualitative analyses of our methods are provided in Appendices~\ref{appsec:comparative_analysis}, \ref{appsec:abstraction_vs_coverage}, and \ref{appsec:criteria_analysis}.

\paragraph{Method Performance Overview}
Across all experiments, \textbf{our advanced prompting methods consistently outperform the basic example-based methods}, with criteria and description variants performing similarly. AUC analysis shows a tight cluster: Criteria (De: 76.3\%, Ex: 76.2\%), Description (Cr: 75.9\%, Ex: 75.6\%), while Summary-Ex (72.0\%) and Examples (69.3\%) trail behind.

\paragraph{Few-shot Scaling Patterns}
Our analysis reveals distinct patterns in average macro-F1 scores as few-shot examples increase. The most striking finding is that \textbf{the standard few-shot (Examples) method exhibits severe performance degradation as the number of few-shot examples increases}. This phenomenon is consistent across all experiments, highlighting the inherent difficulty of our tasks. The Examples method degrades dramatically from 71.5\% at 0-shot to 60.7\% at 100-shot on average. This observation is also true for the Summary-Ex method, which shows similar degradation down to 64.7\%, implying that summarization alone cannot fully address this challenge.

This decline aligns with recent findings that long-context model performance often plateaus or degrades with increased context size \cite{li_long-context_2024, lee-etal-2025-ethic, modarressinolima2025}, a trend likely exacerbated by the complex relational structures and specialized nuances within our dataset.

Remarkably, our proposed methods show robustness to context length, with \textbf{Description-Ex showing steady improvement} from 0-shot (71.5\%) to 100-shot (72.6\%) with minimal fluctuation. \textbf{Criteria-De demonstrates the largest overall gains} from 69.3\% to 72.2\%. In contrast, \textbf{Description-Cr and Criteria-Ex peak at intermediate shot counts}, both at 25 shots with 72.4\% and 71.9\% respectively, before declining to 71.3\% and 70.3\%, indicating performance saturation.

\paragraph{Model Comparison}
Figure~\ref{fig:combined_performance} (Top) indicates that \textbf{Sonnet 3.7 achieves the highest performance overall} with 77\% macro-average F1, showing strong scaling and peak performance at 25 shots. GPT-4o performs consistently until 50 shots before declining, while both Mistral models exhibit clear degradation as shots increase, with Mistral Small peaking early at 10 shots. Haiku 3 shows minimal variation between medium and high shots after an initial performance decline. These findings suggest that \textbf{larger models often benefit from additional exemplars, while smaller models struggle with increased context length.}

\paragraph{Concept Patterns}
The performance across concepts, shown in Figure~\ref{fig:combined_performance} (Bottom), reveals distinct patterns in classification effectiveness and method suitability. Decision Makers consistently achieves the highest F1 scores (reaching 80\%) across all methods, indicating that identifying stakeholders involves more recognizable linguistic patterns than other concepts. Notably, Business Goals and Decision Criteria show similar performance profiles with modest differences between methods, suggesting these concepts share comparable semantic structures in B2B conversations. In contrast, Pain Points exhibits the widest performance spread (55\%-75\% F1) with description-based methods decisively outperforming others, confirming that descriptive context significantly enhances the model's ability to recognize problem-oriented language. Finally, the procedural concept of Decision Making Process uniquely favors Criteria-Ex, showing consistent improvement as shot count increases.

These results demonstrate that \textbf{concept characteristics determine optimal prompting strategies: abstract concepts benefit from descriptive context while systematic concepts require structured criteria.} This underscores the need for concept-specific approaches in B2B sales analysis.

\begin{table}[!t]
\small
\centering
\resizebox{\columnwidth}{!}{%
\begin{tabular}{|l|c|c|c|c|c|c|c|}
\hline
\textbf{Method / $N$ Shots} & \textbf{0} & \textbf{10} & \textbf{25} & \textbf{50} & \textbf{75} & \textbf{100} & \textbf{Avg} \\ \hline
Criteria-Ex & 69.3 & 71.2 & 71.9 & 70.6 & 70.4 & 70.3 & 70.6 \\
Description-Ex & \textbf{71.5} & \textbf{71.9} & \textbf{72.0} & \textbf{72.5} & \textbf{72.3} & \textbf{72.6} & \textbf{72.1} \\
\hline
SC & 71.4 & 69.8 & 67.4 & 64.7 & 63.8 & 62.1 & 66.5 \\
LLMLingua-2 & 46.4 & 49.1 & 48.9 & 49.0 & 46.8 & 47.6 & 48.0 \\
\hline
\end{tabular}%
}
\caption{Comparison with token-level compression methods across few-shot example counts. Mean macro-average F1 scores over all models and concepts.}
\label{tab:token_compression_comparison}
\end{table}

\subsection{Token-Level Compression Comparison}
\label{subsec:token_compression_comparison}

To determine whether our performance gains derive from knowledge quality or simple token reduction, we compare our approach against two extractive baselines at a fixed 50\% compression rate. Unlike our generative methods, these baselines prune original tokens without producing new content: Selective Context (SC)~\citep{li-etal-2023-compressing}, which focuses on example-level compression, and LLMLingua-2~\citep{pan-etal-2024-llmlingua}, which compresses the entire prompt. Table~\ref{tab:token_compression_comparison} presents performance averages across all models and concepts (see Appendix~\ref{appsubsec:compression_details} for per-concept breakdowns).

Consistent with our earlier observations, our methods show improving or stable performance as examples increase, while compression baselines show opposite trends. LLMLingua-2 remains below 50\% throughout, while SC declines substantially from 71.4\% to 62.1\% (a 9.3\% absolute drop), negating the benefit of additional examples.

These trajectories reveal fundamental limitations of these methods. \textbf{Extractive token compression either prunes essential task instructions or yields fragmented examples composed of disconnected salient tokens, failing to provide coherent guidance as example density increases.} LLMLingua-2 fails because it cannot distinguish between compressible content and task-defining elements. SC avoids this by targeting examples specifically. However, its independent token-level pruning yields examples that lack interpretability and relational connectivity, preventing performance gains as example density increases. Moreover, like traditional ICL (Section~\ref{subsec:computation_efficiency}), both methods scale linearly with example count and require training specialized models, introducing significant preprocessing overhead.

In contrast, our methods distill knowledge into interpretable, generalizable patterns through structured distillation. This yields both superior performance and human-readable representations via a single LLM call that does not require any training.

\begin{figure}[t]
  \includegraphics[width=\columnwidth]{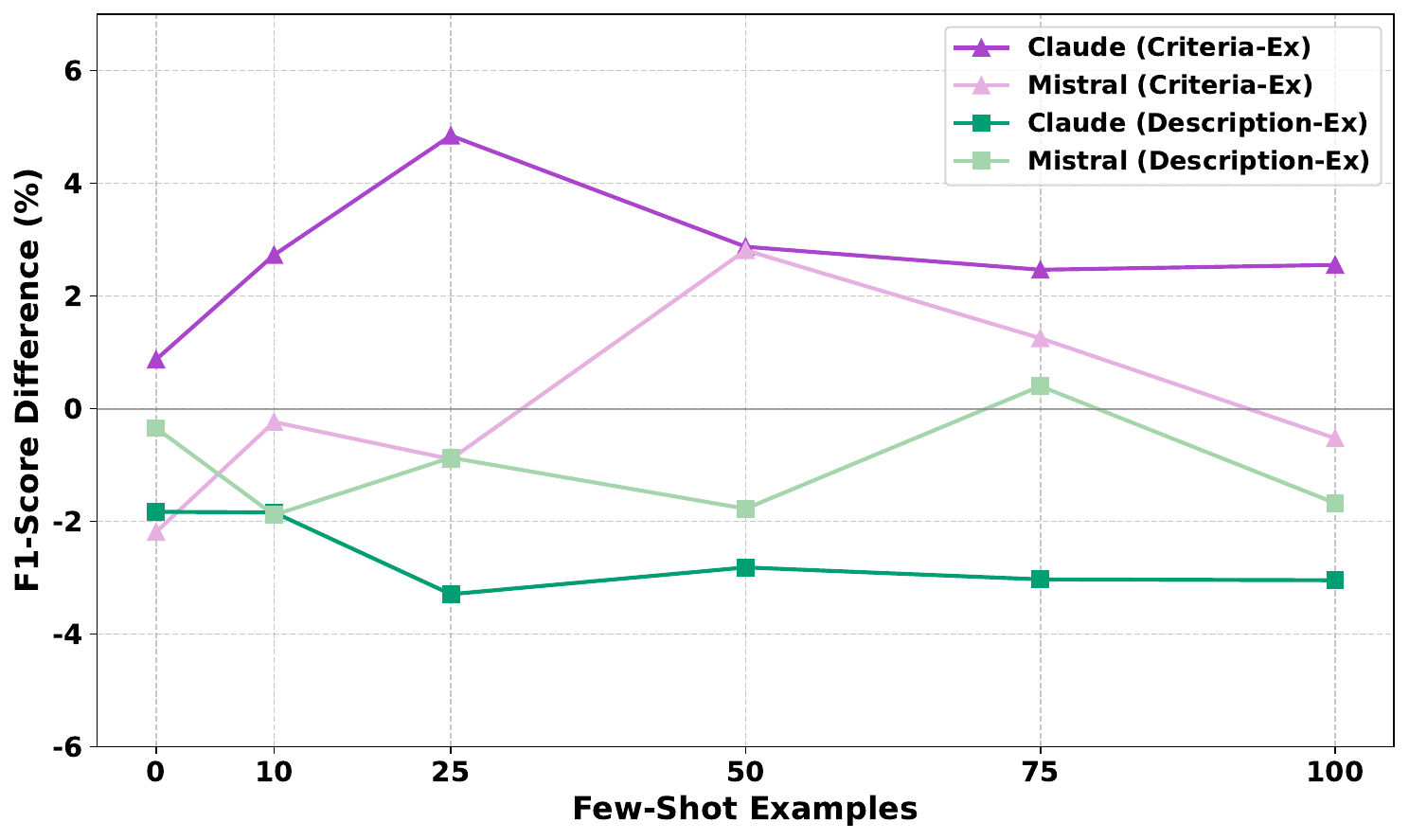}
  \caption{Mean macro-average F1 difference when applying criteria and descriptions generated by larger versus smaller models. Classification was performed by smaller models in all cases. Positive values indicate better performance with larger model-generated content.}
  \label{fig:large_vs_small_performance}
\end{figure}

\subsection{Cross-Model Knowledge Distillation}
\label{subsec:cross_model_distillation}

Unlike standard few-shot learning, our knowledge extraction methods allow us to leverage the generation capabilities of larger, more capable models to enhance the classification performance of smaller models. This approach requires only a single generation step, with minimal time or cost overhead.

We generate criteria and descriptions with the larger models (Sonnet 3.7, Mistral Large) and inject them into the in-context prompts of the smaller models (Haiku 3, Mistral Small) for classification.

Figure~\ref{fig:large_vs_small_performance} illustrates the macro-average F1 difference, averaged over all five concepts, between content generated by the larger models and the original content produced by the smaller models. The results reveal a clear pattern: smaller models consistently improve when using structured criteria from larger models, but perform worse when incorporating descriptions generated by larger models. Mistral models demonstrate lower variance compared to Claude models, although both model families follow the same overall trend. This suggests that \textbf{structured criteria transfer more effectively between models than verbose descriptions}. The rule-based, concise nature of criteria appears more universal than verbose text, which can be highly specific to the source model's generation style, enabling smaller models to better leverage the distilled conceptual structure.

\begin{table}[!t]
\small
\centering
\begin{tabular}{|l|c|c|}
\hline
\textbf{Model / Annotator} & \textbf{Criteria-Ex} & \textbf{Description-Ex} \\ \hline
Claude Sonnet 3.7 & 77.94 & 79.84 \\ \hline
Annotator 1 & \underline{\textbf{80.59}} & \underline{\textbf{81.84}} \\
Annotator 2 & 75.63 & 79.80 \\
Annotator 3 & \underline{79.52} & \underline{80.90} \\
Annotator 4 & \underline{80.30} & \underline{80.50} \\
Annotator 5 & 77.92 & \underline{80.81} \\ \hline
\end{tabular}
\caption{Mean macro-average F1 comparing model-generated knowledge against human-refined versions.}
\label{tab:human_in_the_loop_avg}
\end{table}

\subsection{Human-in-the-Loop Enhancement}
\label{subsec:hitl}

To assess the interpretability and degree of human contribution to our methods, we conducted an experiment involving human annotators. We selected the criteria and descriptions generated by Sonnet 3.7 (our strongest model) from the most effective few-shot size of 25 examples from the first iteration. Five human annotators were then asked to modify the texts generated by Criteria-Ex and Description-Ex for all five concepts. Annotators were given full freedom to revise the generated text by editing, removing, or adding content, while preserving reasonable similarity to the original output.

Table~\ref{tab:human_in_the_loop_avg} compares the original model-generated elements and the human-modified versions in terms of mean macro-average F1 scores across the five tasks. Notably, three annotators improved Criteria-Ex performance and one achieved similar results, while four improved Description-Ex performance, with gains up to 2.65\% and 2\%, respectively. Descriptions proved more conducive to human refinement than criteria, likely because descriptions allow for more flexible reformulation without altering the underlying logic, while criteria modifications can more easily introduce unintended logical conflicts or overlook boundary conditions. Unlike traditional few-shot approaches where classification logic remains opaque, \textbf{our methods provide interpretable artifacts that enable effective human-in-the-loop collaboration, successfully combining LLM strengths with human expertise.} A detailed analysis of human modifications and their impact on performance is provided in Appendix~\ref{appsec:human_modification_example}.

\begin{figure}[!htb]
  \includegraphics[width=\columnwidth]{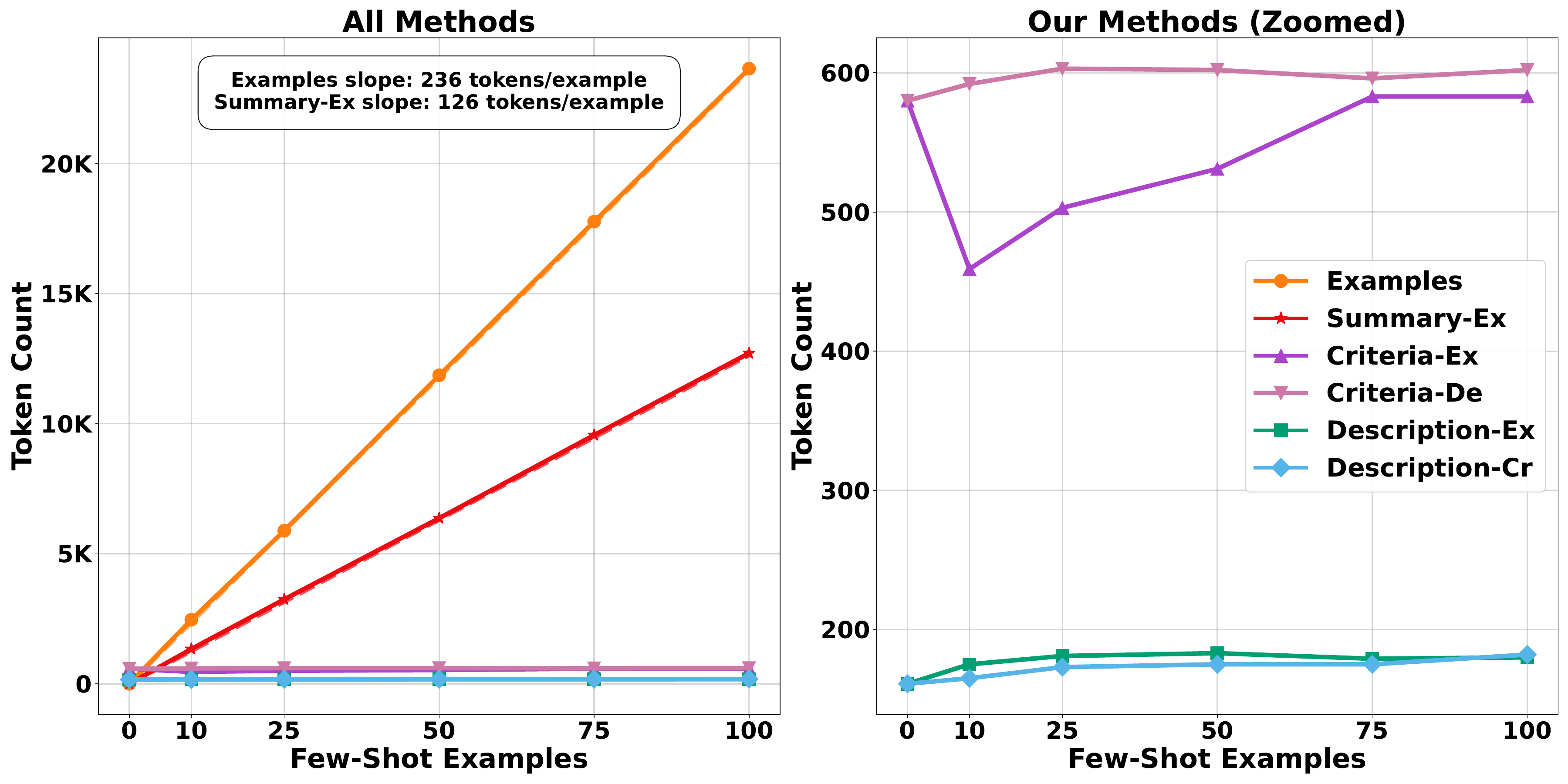}
  \caption{Token count analysis. Left: All methods, showing linear growth for the Examples and Summary methods, while the Criteria and Description methods remain flat. Right: Zoomed view of our methods, revealing minimal variation as the example count increases.}
  \label{fig:token_count_dual}
\end{figure}

\subsection{Computational Efficiency Analysis}
\label{subsec:computation_efficiency}

Figure~\ref{fig:token_count_dual} compares token consumption across methods using the GPT-4o model, averaged over all test sets. Both example-based methods scale linearly with dataset size: Examples at 236 tokens per example and Summary-Ex at 126 tokens per example, reaching 25K and 12.5K tokens, respectively, at 100 examples. In contrast, our methods show minimal correlation with example count: \textbf{Description-based approaches remain under 200 tokens while Criteria-based methods stabilize under 600 tokens regardless of sample size, representing a reduction of up to 99\% in token usage.}

This token efficiency translates directly to faster test-time processing (Figure~\ref{fig:overall_processing_time}), yielding 70\% and 57\% time reductions for criteria and description methods over the Examples approach. Summary-Ex shows moderate efficiency gains, maintaining reasonable performance up to 50 examples before scaling substantially. These results reveal that our proposed methods significantly improve efficiency while preserving performance.

\section{Conclusions}
\label{sec:conclusions}

This work enhances the efficacy of ICL by introducing novel knowledge extraction methods that distill examples into precise task instructions. We contribute a new benchmark spanning five essential B2B concepts, offering a valuable resource for advancing classification in professional contexts. Our dataset reveals the inherent complexity of B2B language understanding, where traditional and complex few-shot methods degrade sharply as context length increases, highlighting the need for more sophisticated approaches.

Through extensive experiments across varied LLMs and few-shot settings, our methods demonstrate superior performance, time efficiency, and cost-effectiveness compared to the standard ICL and various token-level compression techniques. The proposed framework exhibits strong adaptability across different business concepts without requiring domain-specific customization. Crucially, our methods generate interpretable artifacts that enable seamless human-in-the-loop collaboration. This interpretability, combined with our flexible architecture that allows knowledge distillation from powerful to efficient models, opens new pathways for transparent and scalable NLP applications.

\section*{Limitations}

The scope and implications of our research are subject to several limitations.

Although our experimental evaluation is grounded in a carefully curated resource of 50 B2B sales calls, each lasting 30 to 90 minutes and annotated by three independent professional annotators under expert supervision, broader validation across additional B2B corpora remains necessary to fully establish the robustness of our findings. The current scope reflects the resource-intensive nature of expert human annotation, as we prioritize expert-annotated labels over automated alternatives.

Furthermore, our knowledge extraction framework relies on the reasoning capabilities of the underlying LLM to distill accurate criteria from verbose examples. Consequently, our method is susceptible to error propagation if the extraction model hallucinates incorrect rules or misses subtle semantic nuances. While our results show performance gains, the quality of the generated distilled artifacts is upper-bounded by the capability of the model used for distillation.

Finally, our work focuses on binary classification tasks. While our methods can easily be extended to multi-class classification, the extracted knowledge representation would increase linearly with the number of classes, potentially creating efficiency concerns. Furthermore, as the number of classes increases, some may be underrepresented in the sampled examples, limiting the quality of the generated criteria and descriptions. This may compromise generalization capabilities and overall performance in multi-class scenarios.

\begin{comment}
\section*{Acknowledgments}

This document has been adapted
by Steven Bethard, Ryan Cotterell and Rui Yan
from the instructions for earlier ACL and NAACL proceedings, including those for
ACL 2019 by Douwe Kiela and Ivan Vuli\'{c},
NAACL 2019 by Stephanie Lukin and Alla Roskovskaya,
ACL 2018 by Shay Cohen, Kevin Gimpel, and Wei Lu,
NAACL 2018 by Margaret Mitchell and Stephanie Lukin,
Bib\TeX{} suggestions for (NA)ACL 2017/2018 from Jason Eisner,
ACL 2017 by Dan Gildea and Min-Yen Kan,
NAACL 2017 by Margaret Mitchell,
ACL 2012 by Maggie Li and Michael White,
ACL 2010 by Jing-Shin Chang and Philipp Koehn,
ACL 2008 by Johanna D. Moore, Simone Teufel, James Allan, and Sadaoki Furui,
ACL 2005 by Hwee Tou Ng and Kemal Oflazer,
ACL 2002 by Eugene Charniak and Dekang Lin,
and earlier ACL and EACL formats written by several people, including
John Chen, Henry S. Thompson and Donald Walker.
Additional elements were taken from the formatting instructions of the \emph{International Joint Conference on Artificial Intelligence} and the \emph{Conference on Computer Vision and Pattern Recognition}.
\end{comment}

\bibliography{ICL,B2B,B2B_wo_Zotero,GENERAL}

\appendix

The supplementary material is organized as follows: Appendices~\ref{appsec:data_curation_and_guidelines}, \ref{appsec:data_anonymization}, and \ref{appsec:dataset_examples} detail data curation, licensing, ethical governance, anonymization procedures, and representative examples of the dataset. Appendices~\ref{appsec:classification_methodology} and \ref{appsec:knowledge_extraction} describe the classification structure and knowledge extraction methodology. Appendices~\ref{appsec:comparative_analysis}, \ref{appsec:abstraction_vs_coverage}, \ref{appsec:criteria_analysis}, and \ref{appsec:human_modification_example} provide thematic analyses: comparing classification criteria versus descriptions, evaluating generated descriptions through the lens of abstraction versus example coverage, qualitatively assessing generated criteria, and analyzing human-in-the-loop modifications. Finally, Appendix~\ref{appsubsec:average_processing_time} contains average processing times, and Appendix~\ref{appsec:full_grid} presents the comprehensive experimental performance grids.

\section{Data Curation and Usage Guidelines}
\label{appsec:data_curation_and_guidelines}

\subsection{Data Licensing and Intended Use}
The dataset is made available for non-commercial research purposes in NLP. The complete license agreement will be provided as part of the data distribution package on the project website. All existing artifacts were used in accordance with their original intended purposes and licensing terms. Derivatives of this dataset must remain within research contexts to maintain compatibility with the original access conditions.

\subsection{Ethical Approval and Governance}
The data collection protocol underwent internal review by the appropriate legal and scientific governance bodies. This process ensured compliance with organizational ethical guidelines and data protection standards.

%The data collection protocol was reviewed and approved by the Chief Legal Officer and the Chief Scientist of the company, who serve as members of the company's AI governance committee. This approval ensures compliance with internal ethical guidelines and data protection standards.

\subsection{Annotator Profile and Recruitment}
The annotation task was completed by female native speakers of American English. All annotators were recruited through established professional networks and compensated at rates consistent with industry standards for linguistic annotation work in the United States.

\section{Data Anonymization}
\label{appsec:data_anonymization} 
To enable public release, we applied a rigorous anonymization process.\footnote{Experiments confirmed no performance degradation between original and anonymized data.}

A trained annotator identified potentially sensitive information across all snippets, following guidelines covering personal names, organizations, products, locations, contact information, and numeric identifiers. We supplemented manual review with automated heuristics to detect capitalized terms and numeric patterns, and employed Claude Sonnet 3.7 as an additional safety check to identify any overlooked entities.

All identified entities were replaced with fictional alternatives while preserving semantic coherence. For numerical data, we substituted original values with randomized numbers from similar ranges, maintaining local consistency where needed. Professional roles, titles, and percentage values remained unchanged to preserve contextual meaning.\footnote{Our repository provides mappings from entity types to the set of fictional replacements used.}

As a final step, we simplified each snippet through controlled sentence-level rewriting using Claude Sonnet 3.7. We processed entire snippets at once to maintain conversational context, while instructing the model to rewrite individual sentences. We designed a specific prompt directing the model to preserve the semantic content while altering the syntactic structure and word choice. This ensured the text retained its original meaning and conversational flow while no longer resembling the original style or phrasing.

Figure~\ref{fig:simplification_prompt} presents the prompt template used for the transcript simplification process. The input for this prompt is a text snippet, divided into enumerated sentences, that needs to be simplified. The prompt facilitates controlled sentence-level rewriting while preserving the semantic content of conversations. When applying this prompt, the LLM rewrites each line individually, maintaining the conversation's structure and essential meaning. 

\begin{figure}[!htb]
\begin{mdframed}[
  linecolor=promptborder,
  linewidth=1pt,
  backgroundcolor=promptbg,
  innertopmargin=10pt,
  innerbottommargin=10pt,
  innerleftmargin=5pt,
  innerrightmargin=5pt,
  roundcorner=2pt
]
\setlength{\parindent}{0pt}

\begin{minipage}[t]{\dimexpr\columnwidth-12pt\relax}
\footnotesize\color{promptpurple}\fontfamily{zi4}\selectfont
\begin{Verbatim}[breaklines=true, breakanywhere=true, breaksymbol=, 
                 commandchars=\\\{\}]
<instructions>
Please do the following based on the text in <snippet></snippet> tags:
- Rewrite the text in each row separately in a simplified, short, and clear way, maintaining the original phrasing as closely as possible.
- Rewrite the text in each row separately, ensuring that each rewritten row corresponds exactly to the same row in the original snippet. DO NOT skip any rows, and do not combine content from multiple rows into one.
- You should retain all the important details of the transcript.
</instructions>

<snippet>
\{TEXT_SNIPPET\}
</snippet>

<format>
x | SPEAKER_AFFILIATION Rewritten text for row number x
y | SPEAKER_AFFILIATION Rewritten text for row number y
...

<example>
**Original Transcript:**        
1 | [PROSPECT\_A] I have a, I think, a meeting at 10 AM. So I need to... Wait a minute.       
2 | [PROSPECT\_A] So.       
3 | [PROSPECT\_A] Let me see if, let me check with the team. I will need to confirm it with them.        
4 | [SELLER\_A] OK, I mean, the, the, the presentation is due by Friday.

**Rewritten Transcript:**        
1 | [PROSPECT\_A] I have a 10 AM meeting.        
2 | [PROSPECT\_A] So.       
3 | [PROSPECT\_A] let me check and confirm with the team.         
4 | [SELLER\_A] The presentation is due by Friday.
</example>        
</format>
\end{Verbatim}
\end{minipage}
\end{mdframed}
\captionof{figure}{Prompt template used for transcript simplification.}
\label{fig:simplification_prompt}
\end{figure}

\section{Representative Examples of the Dataset}
\label{appsec:dataset_examples}

Table~\ref{tab:call_highlights_examples} presents representative positive snippets from each of the five concepts, highlighting distinct business aspects reflected in the dataset. The bolded portions highlight the most relevant spans for each category. Colors distinguish between prospect (orange) and seller (blue) utterances. These examples, processed and anonymized according to our procedures, reflect the rich diversity and complexity of B2B dialogues.

\begin{table*}[!htbp]
\centering
\begin{adjustbox}{width=\linewidth}
\footnotesize
\setlength{\tabcolsep}{2pt}
\begin{tabular}{|p{1.8cm}|p{16cm}|}
\hline
\textbf{Dataset} & \textbf{Snippet Example} \\
\hline

Business Goals & 
\textsc{\textcolor[HTML]{E69F00}{[PROSPECT\_A]}} Yeah. I'm head of advertising and analytics. Most of our sales are retail. SchistHorizon Networks is our eCommerce star. \textbf{We're trying to increase sales through Omega Operations channel, using Triton Trades for attention and conversions.} GoldenLeaf Enterprises seems to fit our audience: mostly female, 90-30 split, slightly older millennials. So... yeah, that's... 

\textsc{\textcolor[HTML]{0072B2}{[SELLER\_A]}} Okay. Your site looks nice. What inspired this company? 

\textsc{\textcolor[HTML]{E69F00}{[PROSPECT\_A]}} Yeah. We make bottled water and coffee, exploring other options too. Our focus is sustainability. Our source is in Westvale. 

\textsc{\textcolor[HTML]{0072B2}{[SELLER\_A]}} Okay. 

\textsc{\textcolor[HTML]{E69F00}{[PROSPECT\_A]}} Our motto is ``premium by nature''. We don't use chemicals for alkaline water. It's naturally alkaline due to volcanic filtration. That's the core of everything we do. 

\textsc{\textcolor[HTML]{0072B2}{[SELLER\_A]}} Hey, Jean. 

\textsc{\textcolor[HTML]{D55E00}{[PROSPECT\_B]}} Hi there. Sorry I'm late. I was held up on another call but I'm excited to learn about GoldenLeaf Enterprises. \\
\hline

Decision Criteria & 
\textsc{\textcolor[HTML]{0072B2}{[SELLER\_A]}} Yeah, yeah. Based on historical preferences, we want to discuss a ChalkForest Solutions option for you to position ourselves best. This allows them to do it their preferred way. If they change their mind, that's okay. We want you to be prepared for that. I updated Max before our call about our discussions, Val, including your ongoing proof of concept. He's now up to date on our progress. 

\textsc{\textcolor[HTML]{E69F00}{[PROSPECT\_A]}} Your competitors provided four quotes: monthly and three-year contract options. If you can provide a second quote, that's great because you'll be competing on both options, and I can present choices to management. They can choose either option, and this company can offer both. Now it's about what we want. \textbf{We can discuss which company we prefer, focusing on features, support, and functionality.} Do we like the phones? Or not like the phones, things like that? And get into the detailed aspects? 

\textsc{\textcolor[HTML]{56B4E9}{[SELLER\_B]}} Well, I think that from a financial perspective. \\
\hline

Decision Makers & 
\textsc{\textcolor[HTML]{E69F00}{[PROSPECT\_A]}} That's awesome. Very cool. We'd love to see what's involved. Are there any fees for us to use these services? Or is it just? 

\textsc{\textcolor[HTML]{0072B2}{[SELLER\_A]}} No, it's completely complementary to you. LimitlessLogic was created because companies saw employees using it and spending money. The idea was to provide a business experience similar to LimitlessLogic, encouraging personal use. 

\textsc{\textcolor[HTML]{E69F00}{[PROSPECT\_A]}} Yes. 

\textsc{\textcolor[HTML]{0072B2}{[SELLER\_A]}} Yes. That's how this all came about. We used to charge a 10 percent fee before COVID, but we removed it. It's now completely complementary to your organization. We can set it up on LimitlessLogic and VortexVault, customizing programs for your company. 

\textsc{\textcolor[HTML]{E69F00}{[PROSPECT\_A]}} Sounds good. Excellent. \textbf{We'd love to see how to do this, probably involving our people team.} Okay? They would be the ones to roll it out to the company. If you could send that information to me, that would be fantastic. 

\textsc{\textcolor[HTML]{0072B2}{[SELLER\_A]}} I'll send a follow-up email with the information we discussed. Do you have your calendar available to schedule a demo? Are you available next week? What works best for you? \\
\hline

Decision Making Process & 
\textsc{\textcolor[HTML]{0072B2}{[SELLER\_A]}} No problem. Swapping is not an issue. If someone leaves the company, we can swap them immediately and easily. I do it on our end. If an analyst leaves next month, we can add someone else in April without any problem. Kim and I can easily add them and transfer the saved content. If you move teams, we can transfer all your saved content to your replacement. We do this with all clients. We understand people change roles or leave companies. This is not limited. If someone leaves tomorrow and their replacement leaves in two weeks, No problem. We can swap them. We can't allow sharing because it's hard for us to approve. 

\textsc{\textcolor[HTML]{CC79A7}{[SPEAKER\_A]}} Yes. I understand. \textbf{I need to give this feedback to my partners because the three-process was ineffective}. We like the platform and want to continue as explained. But for a user with a laptop, we need to make a console. We can do it, but we're reducing to four laptops. Now, the four-process is still more ineffective. We need to decide if that works for us. \\
\hline

Pain Points & 
\textsc{\textcolor[HTML]{0072B2}{[SELLER\_A]}} What specifically? OIN is failing and it's an issue now. Is there a reason for the urgency? Is it within the next month? 

\textsc{\textcolor[HTML]{E69F00}{[PROSPECT\_A]}} A workshop over the next quarter. I want to solve the password management problem soon. Yeah, I... 

\textsc{\textcolor[HTML]{0072B2}{[SELLER\_A]}} What happens if you don't? Is it just a personal goal? But nothing else? 

\textsc{\textcolor[HTML]{E69F00}{[PROSPECT\_A]}} If solved, clients get a better user experience. \textbf{I have an inefficient account management team. Half my engineers are fixing password problems constantly.} For the business case, It's about efficiency, and user experience. So people can log in and take their training. Currently, people struggle to log in for training. 

\textsc{\textcolor[HTML]{0072B2}{[SELLER\_A]}} How many people use this monthly? How many should log in versus how many actually do? \\
\hline
\end{tabular}
\end{adjustbox}
\caption{Representative positive examples from the \texttt{Call Playbook} Dataset. Relevant spans are bolded.}
\label{tab:call_highlights_examples}
\end{table*}

\section{Classification Methodology}
\label{appsec:classification_methodology}

This section details our approach to classifying B2B conversation snippets, including the classification prompt structure, user intents, and knowledge section formats used in our experiments.

\subsection{Classification Prompt Structure}
\label{appsubsec:classification_prompt}

Our classification system implements a structured prompt template consisting of three main components: (1) a classification objective incorporating the user-provided intent, (2) a knowledge section derived from labeled examples, and (3) the desired structured output format.

Figure~\ref{fig:classification_prompt} presents the base classification prompt template, which remains consistent across all variations. The prompt instructs the model to analyze a text snippet and determine whether it contains evidence of a specific B2B concept (e.g., ``the prospect discusses their business goals''). The user intents used in our experiments are detailed in Section~\ref{appsubsec:user_task_intent}. Variation across prompts arises from the knowledge section, which includes either examples, summarized examples, criteria, or descriptions, as discussed in Section~\ref{appsubsec:knowledge_section}. The prompt concludes with the text snippet to be classified and specifies the expected response format.

\begin{figure}[!htb]
\begin{mdframed}[
  linecolor=promptborder,
  linewidth=1pt,
  backgroundcolor=promptbg,
  innertopmargin=10pt,
  innerbottommargin=10pt,
  innerleftmargin=5pt,
  innerrightmargin=5pt,
  roundcorner=2pt
]
\setlength{\parindent}{0pt}

\begin{minipage}[t]{\dimexpr\columnwidth-12pt\relax}
\footnotesize\color{promptpurple}\fontfamily{zi4}\selectfont
\begin{Verbatim}[breaklines=true, breakanywhere=true, breaksymbol=, 
                 commandchars=\\\{\}]
<instructions>
Analyze the following text snippet and identify whether \{USER\_INTENT\}.

Provide a detailed reasoning for your decision (chain of thoughts) before delivering the final classification.

Label the snippet as either Positive (if \{USER\_INTENT\}) or Negative (if the snippet does not relate or contain relevant information).

\{KNOWLEDGE\_SECTION\}

Base your classification on the objective and the \{KNOWLEDGE_TYPE\} provided above.
</instructions>

<snippet>
\{TEXT_SNIPPET\}
</snippet>

<format>
<rationale> [Your reasoning] </rationale>
<label> [Positive or Negative] </label>
</format>
\end{Verbatim}
\end{minipage}
\end{mdframed}
\captionof{figure}{Base classification prompt template.}
\label{fig:classification_prompt}
\end{figure}

\subsection{User Intents}
\label{appsubsec:user_task_intent}
We conducted our experiments using five distinct classification objectives, each based on a user-provided intent aligned with one of the five B2B conversational concepts:
\begin{compactitem}
\item \textbf{Business Goals:} ``the prospect discusses their business goals in the context of the purchasing process''.
\item \textbf{Decision Criteria:} ``the prospect discusses their decision criteria in the context of the purchasing process''.
\item \textbf{Decision Makers:} ``the prospect mentions the decision makers involved in the purchasing process''.
\item \textbf{Decision Making Process:} ``the prospect discusses their decision-making process regarding the purchase''.
\item \textbf{Pain Points:} ``the prospect discusses their pain points in the context of the purchasing process''.
\end{compactitem}

These intents were inserted into the classification prompt template at the \texttt{\{USER\_INTENT\}} placeholder shown in Figure~\ref{fig:classification_prompt}. 

The intents are intentionally high-level, reflecting the typical level of specificity provided by users in real-world applications. This level of abstraction in such intents is common, as users may not always possess the domain expertise or technical vocabulary to formulate precise classification parameters. This limitation further motivates our approach of augmenting user-provided intents with knowledge derived from labeled examples.

\subsection{Knowledge Section Formats}
\label{appsubsec:knowledge_section}

As described in Section~\ref{appsubsec:classification_prompt}, we implemented three distinct formats for the \texttt{\{KNOWLEDGE\_SECTION\}} placeholder of the classification prompt: the Examples format (also used by Summary-Ex) shown in Figure~\ref{fig:examples_knowledge_section}, the Criteria format shown in Figure~\ref{fig:criteria_knowledge_section}, and the Description format shown in Figure~\ref{fig:description_knowledge_section}. 

For all formats, positive examples or criteria precede negative ones, as early experiments confirmed that this ordering produced superior performance.

\begin{figure}[H]
\begin{mdframed}[
  linecolor=promptborder,
  linewidth=1pt,
  backgroundcolor=promptbg,
  innertopmargin=10pt,
  innerbottommargin=10pt,
  innerleftmargin=5pt,
  innerrightmargin=5pt,
  roundcorner=2pt
]
\setlength{\parindent}{0pt}

\begin{minipage}[t]{\dimexpr\columnwidth-12pt\relax}
\footnotesize\color{promptpurple}\fontfamily{zi4}\selectfont
\begin{Verbatim}[breaklines=true, breakanywhere=true, breaksymbol=, 
                 commandchars=\\\{\}]
Below is a list of positive examples that would indicate that the objective is present in the text snippet:
<positive_examples>
\{POSITIVE\_EXAMPLES\}
</positive_examples>

Below is a list of negative examples that would not indicate that the objective is present in the text snippet:
<negative_examples>
\{NEGATIVE\_EXAMPLES\}
</negative_examples>
\end{Verbatim}
\end{minipage}
\end{mdframed}
\captionof{figure}{Example-based knowledge guidance format that uses a direct few-shot approach with labeled examples from the dataset.}
\label{fig:examples_knowledge_section}
\end{figure}

\begin{figure}[H]
\begin{mdframed}[
  linecolor=promptborder,
  linewidth=1pt,
  backgroundcolor=promptbg,
  innertopmargin=10pt,
  innerbottommargin=10pt,
  innerleftmargin=5pt,
  innerrightmargin=5pt,
  roundcorner=2pt
]
\setlength{\parindent}{0pt}

\begin{minipage}[t]{\dimexpr\columnwidth-12pt\relax}
\footnotesize\color{promptpurple}\fontfamily{zi4}\selectfont
\begin{Verbatim}[breaklines=true, breakanywhere=true, breaksymbol=, 
                 commandchars=\\\{\}]
Below is a list of positive criteria that would indicate that the objective is present in the text snippet:
<positive_criteria>
\{POSITIVE\_CRITERIA\}
</positive_criteria>

Below is a list of negative criteria that would not indicate that the objective is present in the text snippet:
<negative_criteria>
\{NEGATIVE\_CRITERIA\}
</negative_criteria>
\end{Verbatim}
\end{minipage}
\end{mdframed}
\captionof{figure}{Criteria-based knowledge guidance format where an LLM distills labeled examples into explicit classification criteria.}
\label{fig:criteria_knowledge_section}
\end{figure}

\begin{figure}[H]
\begin{mdframed}[
  linecolor=promptborder,
  linewidth=1pt,
  backgroundcolor=promptbg,
  innertopmargin=10pt,
  innerbottommargin=10pt,
  innerleftmargin=5pt,
  innerrightmargin=5pt,
  roundcorner=2pt
]
\setlength{\parindent}{0pt}

\begin{minipage}[t]{\dimexpr\columnwidth-12pt\relax}
\footnotesize\color{promptpurple}\fontfamily{zi4}\selectfont
\begin{Verbatim}[breaklines=true, breakanywhere=true, breaksymbol=, 
                 commandchars=\\\{\}]
Below is a detailed description of the classification task:
<description>
\{DESCRIPTION\}
</description>
\end{Verbatim}
\end{minipage}
\end{mdframed}
\captionof{figure}{Description-based knowledge guidance format where an LLM generates a comprehensive explanation of the classification task based on labeled examples.}
\label{fig:description_knowledge_section}
\end{figure}

\section{Knowledge Extraction Process}
\label{appsec:knowledge_extraction}

Our approach transforms raw labeled examples into condensed, structured knowledge representations using LLMs. This section introduces the three types of knowledge representations used in our prompts: summaries, criteria, and descriptions.
For each type, we describe the corresponding prompt design and the process used to generate it from labeled examples. Each representation was then inserted into one of the knowledge section formats described in Appendix~\ref{appsubsec:knowledge_section}.

\subsection{Summary Generation}
\label{appsubsec:summary_generation}

To generate summarized examples for the Summary-Ex method, we applied a text summarization process using Claude Sonnet 3.7 that compresses the original labeled snippets while preserving their essential meaning, speaker structure, and conversational flow. Figure~\ref{fig:summary_prompt} shows the prompt template used to generate the summaries.

This method aims to reduce overall prompt length while preserving essential business information and discourse patterns, potentially improving the model's focus on relevant content.

The resulting summaries replace the full-length examples in the Examples format (Figure~\ref{fig:examples_knowledge_section}).

\begin{figure}[!htb]
\begin{mdframed}[
  linecolor=promptborder,
  linewidth=1pt,
  backgroundcolor=promptbg,
  innertopmargin=10pt,
  innerbottommargin=10pt,
  innerleftmargin=5pt,
  innerrightmargin=5pt,
  roundcorner=2pt
]
\setlength{\parindent}{0pt}

\begin{minipage}[t]{\dimexpr\columnwidth-12pt\relax}
\footnotesize\color{promptpurple}\fontfamily{zi4}\selectfont
\begin{Verbatim}[breaklines=true, breakanywhere=true, breaksymbol=, 
                 commandchars=\\\{\}]
<instructions>
Analyze the following B2B call text snippet and create a simplified, concise version of the original text.
Preserve the original format and maintain all speaker affiliations exactly as they appear in the source.
Focus on removing redundant information and filler words while keeping all discussed content, focusing on main business topics.
Keep the essential structure and flow of the conversation intact.

Create a condensed version that captures what was discussed without changing the text format or speaker affiliations.
The summary should include 3-5 sentences at most.
</instructions>

<snippet>
\{TEXT_SNIPPET\}
</snippet>

<format>
Your answer must be in the following format:
<summary>
[Simplified version of the original text]
</summary>
</format>
\end{Verbatim}
\end{minipage}
\end{mdframed}
\captionof{figure}{Summary generation prompt template.}
\label{fig:summary_prompt}
\end{figure}

\subsection{Criteria Generation}
\label{appsubsec:criteria_generation}

To generate classification criteria, we employed two variants. The first variant, Criteria-Ex, derives criteria directly from examples, while the second variant, Criteria-De, derives criteria from an existing task description (generated by Description-Ex). For this purpose, we employed a prompt template that guides the relevant model to generate two lists: positive criteria that indicate the presence of the target concept and negative criteria that indicate its absence. Figure~\ref{fig:criteria_prompt} shows the prompt template used for this purpose.

In both variants, the prompt emphasizes the need for general, clear, and concise criteria that can be applied to any text snippet. In the Criteria-Ex variant, the model is further instructed to base each criterion on patterns observed in at least two of the provided examples. For zero-shot prompting, we omit the \texttt{\{KNOWLEDGE\_SECTION\}} and generate the criteria based solely on the user intent.

Both variants follow the same knowledge section formats outlined in Appendix~\ref{appsubsec:knowledge_section}, as illustrated in the prompt template shown in Figure~\ref{fig:criteria_prompt}, adapting them to the specific requirements of criteria generation by incorporating the appropriate content into the \texttt{\{KNOWLEDGE\_SECTION\}} placeholder, where the \texttt{\{KNOWLEDGE\_TYPE\}} can be either ``examples'' or ``description'' depending on the specific variant being employed.

The resulting criteria are then inserted into the knowledge section embedded within the classification prompt, using the criteria variant (Figure~\ref{fig:criteria_knowledge_section}).

\begin{figure}[!htb]
\begin{mdframed}[
  linecolor=promptborder,
  linewidth=1pt,
  backgroundcolor=promptbg,
  innertopmargin=10pt,
  innerbottommargin=10pt,
  innerleftmargin=5pt,
  innerrightmargin=5pt,
  roundcorner=2pt
]
\setlength{\parindent}{0pt}

\begin{minipage}[t]{\dimexpr\columnwidth-12pt\relax}
\footnotesize\color{promptpurple}\fontfamily{zi4}\selectfont
\begin{Verbatim}[breaklines=true, breakanywhere=true, breaksymbol=, 
                 commandchars=\\\{\}]
<instructions>
You are tasked with annotating text snippets.

The end-goal task is to analyze text snippets and determine whether \{USER\_INTENT\}.

Your task is to generate two lists:
A list of positive criteria: Conditions that indicate the presence of the objective.

A list of negative criteria: Conditions that indicate the absence of the objective.

\{KNOWLEDGE\_SECTION\}

Base your criteria on the objective and the \{KNOWLEDGE\_TYPE\} provided above.
The criteria should be as general as possible and should be applicable to any text snippet.
The criteria should be clear and concise.
Each list of criteria should include at least five criteria and no more than ten criteria.
Each criterion should be self-explanatory and not require an example.
[For Criteria-Ex: Each criterion should be based on at least two of the examples provided above.]
</instructions>

<format>
Your answer must be in the following format:
<criteria>
<positive>
Criterion 1: [Criterion 1]
Criterion 2: [Criterion 2]
...
</positive>
<negative>
Criterion 1: [Criterion 1]
Criterion 2: [Criterion 2]
...
</negative>
</criteria>

[Example format omitted for brevity]
</format>
\end{Verbatim}
\end{minipage}
\end{mdframed}
\captionof{figure}{Criteria generation prompt template.}
\label{fig:criteria_prompt}
\end{figure}

\subsection{Description Generation}
\label{appsubsec:description_generation}

To generate task descriptions, we similarly employed two variants. The first variant, Description-Ex, derives a description directly from examples, while the second variant, Description-Cr, derives a description from existing criteria (generated by Criteria-Ex). To this end, we employed a prompt template that guides the relevant model to produce a detailed description of the classification task. Figure~\ref{fig:description_prompt} shows the prompt template used for description generation.

In both variants, the prompt instructs the model to generate a description of the classification task that facilitates effective recognition of the target concept in conversational text. The instructions emphasize the importance of generating descriptions that strike a balance between comprehensiveness and brevity, ensuring they can be applied consistently across diverse text samples. For zero-shot prompting, we omit the \texttt{\{KNOWLEDGE\_SECTION\}} and generate the description based solely on the user intent.

Both variants utilize the knowledge section formats outlined in Section~\ref{appsubsec:knowledge_section}, as illustrated in the prompt template shown in Figure~\ref{fig:description_prompt}, adapting them to the specific requirements of description generation by incorporating the appropriate content into the \texttt{\{KNOWLEDGE\_SECTION\}} placeholder, where the \texttt{\{KNOWLEDGE\_TYPE\}} can be either ``examples'' or ``criteria'' depending on the specific variant being employed.

The resulting description is then embedded within the classification prompt, using the description variant (Figure~\ref{fig:description_knowledge_section}).

\begin{figure}[!htb]
\begin{mdframed}[
  linecolor=promptborder,
  linewidth=1pt,
  backgroundcolor=promptbg,
  innertopmargin=10pt,
  innerbottommargin=10pt,
  innerleftmargin=5pt,
  innerrightmargin=5pt,
  roundcorner=2pt
]
\setlength{\parindent}{0pt}

\begin{minipage}[t]{\dimexpr\columnwidth-12pt\relax}
\footnotesize\color{promptpurple}\fontfamily{zi4}\selectfont
\begin{Verbatim}[breaklines=true, breakanywhere=true, breaksymbol=, 
commandchars=\\\{\}]
<instructions>
You are tasked with annotating text snippets.

The end-goal task is to analyze text snippets and determine whether \{USER\_INTENT\}.

Your task is to generate a detailed description of the classification task allowing for the identification of the objective in text snippets.

\{KNOWLEDGE\_SECTION\}

Base your description on the objective and the \{KNOWLEDGE\_TYPE\} provided above.
The description should be as general as possible and should be applicable to any text snippet.
The description should be clear and concise.
</instructions>

<format>
Your answer must be in the following format:
<description>
[Your description]
</description>

[Example format omitted for brevity]
</format>
\end{Verbatim}
\end{minipage}
\end{mdframed}
\captionof{figure}{Description generation prompt template.}
\label{fig:description_prompt}
\end{figure}

\section{Comparative Analysis of the Criteria and Description Methods}
\label{appsec:comparative_analysis}

This appendix presents a comparative analysis of the four criteria and description methods. Tables~\ref{tab:examplified_criteria} and~\ref{tab:examplified_descriptions} show the texts generated by Claude Sonnet 3.7 using 25 few-shot examples from the first iteration, applied to the Business Goals concept.

To illustrate how these task instructions function in practice, we analyze their application to two positive and two negative representative examples:

\begin{itemize}
\item \textcolor{Blue}{Positive 1:} A conversation in which the prospect articulates how they view their customer service channel as a revenue opportunity, referencing specific metrics such as call handling times and outlining their business goals for transforming it into a more marketing-oriented function.

\item \textcolor{teal}{Positive 2:} A conversation in which the prospect describes content management challenges, specifically their need to maintain consistency across product documentation and how centralized updates would improve operational efficiency.

\item \textcolor{red}{Negative 1:} A conversation in which the participants focus entirely on small talk about an earthquake and personal topics, with no articulation of business objectives.

\item \textcolor{brown}{Negative 2:} A conversation in which the participants focus solely on contract terms and technical settings between sellers, without any expression of the prospect's business goals or strategic needs.
\end{itemize}

The side-by-side format in both tables separates the instructional content (criteria definitions or descriptive guidance) from the illustrative examples, making it easier to see how these examples influenced the generation of the criteria and descriptions, and showing the different approaches to identifying business objectives in sales conversations.

\begin{table*}[htbp]
\footnotesize
\begin{tabular}{|p{0.04\textwidth}|p{0.38\textwidth}|p{0.50\textwidth}|}
\hline
\multicolumn{3}{|c|}{\textbf{Criteria-Ex: Positive}} \\
\hline
\textbf{ID} & \textbf{Criteria} & \textbf{Illustrative Quotes} \\
\hline
Crit. 1 & The prospect explicitly states their business objectives or goals that they hope to achieve through the purchase or implementation of the product/service & \textcolor{Blue}{\textbf{``We want that channel to be more marketing-minded and product-focused rather than service-focused.''}} \\
\hline
Crit. 2 & The prospect discusses how the product/service would integrate with or improve their existing business processes, workflows, or operations & \textcolor{teal}{\textbf{``We wonder if it's possible to maintain all this data in a single place... so updating one place automatically updates all documents.''}} \\
\hline
Crit. 3 & The prospect explains specific business challenges or pain points they are trying to solve through the purchasing decision & \textcolor{teal}{\textbf{``When we change one document, it must be changed in all documents. That's our current challenge.''}} \\
\hline
Crit. 6 & The prospect shares information about their business model, operational structure, or customer relationships in the context of how the purchase would impact these areas & \textcolor{Blue}{\textbf{``Most calls are about order tracking, which we can address with self-service... When a rep is on a product inquiry call, about six other customers are waiting for service.''}} \\
\hline
\multicolumn{3}{|c|}{\textbf{Criteria-Ex: Negative}} \\
\hline
Crit. 2 & The prospect discusses pricing, contracts, or payment terms without relating them to broader business goals or outcomes & \textcolor{brown}{\textbf{``They're also in a two-year contract. If they add seats, they'd pay for six years, right?''}} \\
\hline
Crit. 5 & The conversation consists primarily of small talk, introductions, or unrelated topics that don't touch on business objectives & \textcolor{red}{\textbf{``Everyone's talking about the Ironwood earthquake over the last 25 minutes here on the Corswick. It was in Ivorycliff.''}} \\
\hline
Crit. 7 & The conversation is dominated by the seller explaining their offering without the prospect articulating how it connects to their business objectives & \textcolor{brown}{\textbf{``I'll ask if they want 20 seats or remind them account sharing isn't allowed. We can let them out of the contract. If they stay, they need more users.''}} \\
\hline
\multicolumn{3}{|c|}{\textbf{Criteria-De: Positive}} \\
\hline
Crit. 2 & The prospect describes specific problems or pain points in their current business operations that they are looking to solve through the purchasing decision & \textcolor{teal}{\textbf{``Between reports, about 24 percent of the data is common... When we change one document, it must be changed in all documents.''}} \\
\hline
Crit. 3 & The prospect articulates measurable targets, metrics, or key performance indicators they aim to improve through the acquisition of a product or service & \textcolor{Blue}{\textbf{``A customer inquiring about a product takes 6 to 20 minutes to close. A customer service call takes about two minutes... When a rep is on a product inquiry call, about six other customers are waiting for service.''}} \\
\hline
Crit. 6 & The prospect outlines specific operational efficiencies, cost savings, or productivity improvements they expect to gain from the purchase & \textcolor{teal}{\textbf{``One advantage we see is building blocks, as you mentioned. Updating one block should update all reports.''}} \\
\hline
Crit. 7 & The prospect connects features or capabilities of the product/service directly to their business needs or organizational priorities & \textcolor{Blue}{\textbf{``We see an opportunity because we have a wide range of products that require education about fit. For our women's assortment, we have about seven different fits, multiplied by 25 fabric types, colors, and washes. It's complex.''}} \\
\hline
\multicolumn{3}{|c|}{\textbf{Criteria-De: Negative}} \\
\hline
Crit. 2 & The conversation consists primarily of the seller explaining potential benefits without the prospect articulating their own business goals or needs & \textcolor{brown}{\textbf{``I'll ask if they want 20 seats or remind them account sharing isn't allowed. We can let them out of the contract. If they stay, they need more users.''}} \\
\hline
Crit. 4 & The text contains only small talk, pleasantries, or relationship-building conversation unrelated to business goals or purchasing decisions & \textcolor{red}{\textbf{``You hear me? Yeah, hi, Kenzie. How are you?... I'm in Quorvath... My family is on the Corswick... My sister is flying to Torrengard tonight. She plays lacrosse.''}} \\
\hline
Crit. 5 & The prospect discusses only pricing, contract terms, or payment options without relating these to their business objectives or expected outcomes & \textcolor{brown}{\textbf{``They complained, so we set them to two for nine users. They're on two now.''}} \\
\hline
\end{tabular}
\caption{Comparative Analysis of Criteria-Based Instructions: Highlighted quotes illustrate how positive examples (blue/teal) inform the positive criteria and negative examples (red/brown) inform the negative criteria. The table demonstrates how structured criteria from both the Criteria-Ex and Criteria-De approaches capture different aspects of the Business Goals concept.}
\label{tab:examplified_criteria}
\end{table*}

\begin{table*}[!htb]
\footnotesize
\begin{tabular}{|p{0.45\textwidth}|p{0.47\textwidth}|}
\hline
\multicolumn{2}{|c|}{\textbf{Description-Ex}} \\
\hline
\textbf{Description} & \textbf{Illustrative Quotes} \\
\hline
The task is to determine whether any part of the text shows the prospect discussing their business goals, objectives, or desired outcomes in relation to a purchasing decision or process & \textcolor{Blue}{\textbf{``I definitely see it as an opportunity... I see it as a sales channel.''}} \newline
\textcolor{teal}{\textbf{``We have multiple products and need to prepare reports for each one... That's our current challenge.''}} \\
\hline
This includes when prospects explain what they want to achieve with a product/service. Prospects may describe their organization's strategic aims or outline operational needs & \textcolor{Blue}{\textbf{``We want that channel to be more marketing-minded and product-focused rather than service-focused.''}} \newline
\textcolor{teal}{\textbf{``So updating one place automatically updates all documents, like for a plant manufacturing 6 products.''}} \\
\hline
Look for instances where prospects articulate their vision, priorities, requirements, or expected benefits from implementing a solution & \textcolor{Blue}{\textbf{``We see an opportunity because we have a wide range of products that require education about fit.''}} \newline
\textcolor{teal}{\textbf{``One advantage we see is building blocks, as you mentioned. Updating one block should update all reports.''}} \\
\hline
This may involve discussions about improving processes, solving problems, or enhancing efficiency & \textcolor{teal}{\textbf{``When we change one document, it must be changed in all documents. That's our current challenge... We wonder if it's possible to maintain all this data in a single place.''}} \newline
\textcolor{Blue}{\textbf{``Most calls are about order tracking, which we can address with self-service and more resources.''}} \\
\hline
Exclude general small talk or discussions where only the seller is talking without the prospect articulating their own business objectives & \textcolor{red}{\textbf{``Everyone's talking about the Ironwood earthquake over the last 25 minutes here on the Corswick.''}} \newline
\textcolor{brown}{\textbf{``I'll ask if they want 20 seats or remind them account sharing isn't allowed. We can let them out of the contract. If they stay, they need more users.''}} \\
\hline
\multicolumn{2}{|c|}{\textbf{Description-Cr}} \\
\hline
The task is to determine whether any part of the text shows the prospect articulating their business goals, objectives, or desired outcomes in relation to a potential purchase or implementation & \textcolor{Blue}{\textbf{``I definitely see it as an opportunity... I see it as a sales channel. We want that channel to be more marketing-minded and product-focused rather than service-focused.''}} \newline
\textcolor{teal}{\textbf{``We wonder if it's possible to maintain all this data in a single place... Updating one block should update all reports.''}} \\
\hline
Look for instances where the prospect connects the purchase decision to measurable business outcomes, ROI expectations, or organizational growth plans. The objective is present when prospects share insights about their business model or operational structure & \textcolor{Blue}{\textbf{``A customer inquiring about a product takes 6 to 20 minutes to close. A customer service call takes about two minutes... When a rep is on a product inquiry call, about six other customers are waiting for service.''}} \newline
\textcolor{Blue}{\textbf{``We have a wide range of products that require education about fit. For our women's assortment, we have about seven different fits, multiplied by 25 fabric types, colors, and washes. It's complex.''}} \\
\hline
Exclude conversations that focus solely on technical specifications, pricing details, or administrative aspects without connection to broader business goals. Also exclude instances when the conversation is dominated by the seller without the prospect articulating how the offering aligns with their business objectives, or contains only pleasantries & \textcolor{brown}{\textbf{``They're also in a two-year contract. If they add seats, they'd pay for six years, right?... They complained, so we set them to two for nine users.''}} \newline
\textcolor{brown}{\textbf{``I'll ask if they want 20 seats or remind them account sharing isn't allowed. We can let them out of the contract.''}} \newline
\textcolor{red}{\textbf{``You hear me? Yeah, hi, Kenzie. How are you?... Everyone's talking about the Ironwood earthquake.''}} \\
\hline
\end{tabular}
\caption{Comparative Analysis of Description-Based Instructions: Highlighted quotes illustrate how positive examples (blue/teal) inform the positive guidance and negative examples (red/brown) inform the exclusion guidance. The table demonstrates how contextual descriptions from both Description-Ex and Description-Cr approaches provide holistic frameworks for the Business Goals concept.}
\label{tab:examplified_descriptions}
\end{table*}

Our analysis reveals several key distinctions and similarities between the four instructional approaches:

\textbf{Structural Differences}: The criteria-based approaches (Criteria-Ex and Criteria-De) provide discrete, categorical guidelines that annotators can apply systematically. In contrast, the narrative descriptions (Description-Ex and Description-Cr) offer more contextual guidance and flow more naturally, potentially making them more accessible to non-expert annotators.

\textbf{Emphasis Variations}: Criteria-Ex emphasizes how the prospect relates a product or service to their business context, while Criteria-De focuses more on the nature of the business objectives themselves. Description-Ex highlights conversational indicators of business goals, whereas Description-Cr emphasizes measurable outcomes and operational insights.

\textbf{Complementary Coverage}: All four approaches effectively identify the positive examples as containing business objectives, but through different analytical lenses. For instance, Positive 1 is recognized through explicit goal statements in Criteria-Ex but through measurable metrics in Criteria-De.

\textbf{Negative Case Identification}: Each approach successfully flags the negative examples as lacking business objective articulation, but with different emphasis: Negative 1 is identified primarily through its small-talk nature, while Negative 2 is flagged for its focus on administrative details without business context.

\textbf{Granularity vs. Holistic Assessment}: The criteria-based approaches offer more granular analytical points, potentially supporting more consistent annotation across different raters. The narrative descriptions provide a more holistic framework that may better capture contextual nuances in articulating business objectives.

This analysis demonstrates how our task instructions guide annotators through different analytical pathways. While criteria-based methods enable systematic decomposition of conversational elements, narrative descriptions encourage comprehensive evaluation of speaker intent. These findings inform the design of robust annotation protocols that balance analytical precision with contextual sensitivity in conversational analysis.

\begin{figure}[!ht]
  \includegraphics[width=\columnwidth]{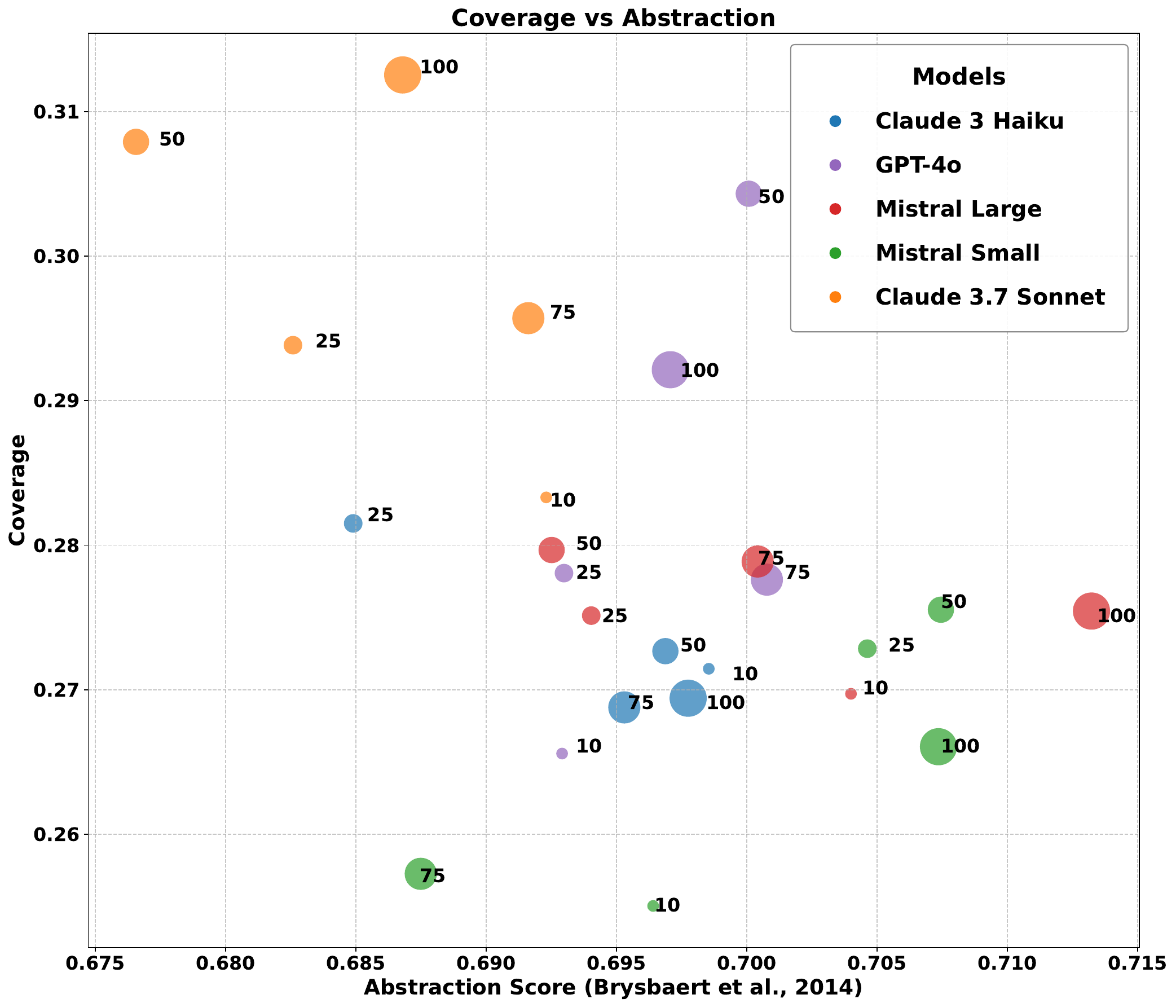}
  \caption{Trade-off between abstraction and coverage in task descriptions. Point size indicates the number of examples used. Higher abstraction scores reflect more abstract descriptions, while higher coverage indicates a better representation of examples.}
  \label{fig:coverage_vs_abstraction}
\end{figure}

\section{Abstraction vs. Coverage in LLM-Generated Descriptions}
\label{appsec:abstraction_vs_coverage}

In Figure~\ref{fig:coverage_vs_abstraction}, we examine how LLMs balance coverage and abstraction through the descriptions generated by Description-Ex. Coverage is measured as the cosine similarity between model-generated descriptions and their source examples using \textit{all-MiniLM-L6-v2} embeddings, while abstraction is quantified using the concreteness ratings from \citet{brysbaert2014concreteness}. Each point represents a model-dataset pairing with varying few-shot counts (indicated by point size), revealing how these properties interact across different dimensions.

The figure demonstrates a clear trade-off between these properties. Sonnet 3.7 achieves the highest coverage scores, particularly with larger example sets, but operates at lower abstraction levels. Conversely, both Mistral models demonstrate superior abstraction capabilities while exhibiting reduced coverage of the original examples. 

This pattern highlights an inherent tension: \textbf{descriptions that closely mirror their source examples tend to be more concrete, while more abstract descriptions capture fewer specific details from the training data.} GPT-4o presents a more balanced approach, achieving good abstraction while retaining reasonable coverage, especially when provided with moderate to large example sets.

Interestingly, while we expected more examples to increase abstraction and reduce coverage, only Mistral models follow this pattern. Sonnet 3.7 exhibits the opposite trend, and Haiku 3 shows no clear correlation, suggesting that these LLMs differ in their learning approaches.

\section{Qualitative Analysis of Generated Classification Criteria}
\label{appsec:criteria_analysis}

To better understand the characteristics of the generated criteria for our B2B concepts, we conducted a systematic analysis of more than 3,000 criteria produced across all experimental conditions. We examined several key linguistic properties for each criterion:
\begin{compactitem}
\item \textbf{Number of criteria}: The average number of positive and negative criteria generated per class, reflecting the model’s ability to articulate classification rules.
\item \textbf{Abstraction}: Computed using established concreteness ratings from the \citet{brysbaert2014concreteness} lexical database, with higher scores indicating more abstract language.
\item \textbf{Business indicators}: Identified using pattern matching against a comprehensive lexicon of business value terminology (e.g., ROI, customer retention, market share).
\item \textbf{Implementation focus}: Detected through the presence of technical and procedural terminology related to system deployment and execution processes.
\item \textbf{Solution orientation}: Assessed based on the presence of solution-focused verbs and outcome-oriented language.
\item \textbf{Conditional logic}: Tracked through business-relevant conditional statements and logical constructions.
\end{compactitem}

\begin{table}[!htbp]
\centering
\begin{adjustbox}{width=\linewidth}
\begin{tabular}{|l|c|c|}
\hline
\textbf{Property} & \textbf{Criteria-Ex} & \textbf{Criteria-De} \\ \hline
Positive criteria & 6.0 & 6.5 \\
Negative criteria & 6.0 & 6.5 \\
Abstraction (0-1 scale) & 0.69 & 0.69 \\
Business value indicators & 64.4\% & 58.7\% \\
Implementation details & 59.0\% & 49.0\% \\
Solution-focused language & 29.1\% & 24.8\% \\
Conditional business logic & 33.9\% & 38.1\% \\
Quantifiable metrics & 0.1\% & 0.0\% \\ \hline
\end{tabular}
\end{adjustbox}
\caption{Linguistic characteristics of generated classification criteria.}
\label{tab:criteria_linguistics}
\end{table}

Table~\ref{tab:criteria_linguistics} presents the differences between criteria derived from examples versus those derived from descriptions, revealing distinct linguistic patterns. Our analysis reveals that Criteria-Ex exhibits higher rates of business value references (+5.7\%) and implementation details (+10.0\%) compared to Criteria-De. Both approaches yield identical high abstraction scores (0.69) and maintain comparable numbers of positive and negative criteria, with Criteria-De generating slightly more criteria per sample.

A notable finding is the near absence of quantifiable metrics across all criteria, suggesting that models prioritize qualitative over quantitative reasoning when establishing classification guidelines. Additionally, both approaches demonstrate relatively low usage of conditional logic, although Criteria-De employs conditional statements more frequently (38.1\%) than Criteria-Ex (33.9\%). This pattern indicates that the generated criteria tend to favor declarative statements over conditional or logical formulations.

\begin{table*}[!ht]
\centering
\small
\begin{tabular}{|p{2.5cm}|p{4cm}|p{4cm}|p{4cm}|}
\hline
\textbf{Criterion} & \textbf{Claude Sonnet 3.7} & \textbf{Annotator 1} \newline \textbf{(+9.8\% F1)} & \textbf{Annotator 2} \newline \textbf{(-1.3\% F1)} \\
\hline
\textbf{Positive 1} & The prospect explicitly describes challenges, frustrations, or difficulties they are experiencing with their current purchasing process, systems, or vendors. & The prospect explicitly describes challenges, frustrations, or difficulties they are experiencing with their current \textbf{solutions}, systems, or vendors. & The \textbf{potential customer openly discusses struggles}, frustrations, or issues they are encountering with their current purchasing process, systems, or \textbf{suppliers}. \\
\hline
\textbf{Positive 2} & The prospect mentions inefficiencies, wasted time, or resource constraints that are directly impacting their ability to make purchasing decisions or implement solutions. & The prospect mentions inefficiencies or \textbf{wasted time}. & The \textbf{potential customer} points out inefficiencies, wasted time, or resource limitations that are \textbf{directly affecting their capability to make purchase decisions or implement solutions}. \\
\hline
\textbf{Negative 1} & The prospect discusses general business information, company background, or role descriptions without mentioning any challenges or difficulties related to purchasing processes. & The prospect discusses general business information, company background, or role descriptions without mentioning any \textbf{challenges or difficulties}. & The \textbf{potential customer} discusses general business \textbf{data}, company \textbf{history}, or role descriptions without mentioning any \textbf{struggles or issues related to purchasing processes}. \\
\hline
\textbf{Negative 6} & The conversation focuses primarily on logistics, scheduling, or administrative details of the current meeting rather than business challenges. & \textit{(No change)} & The conversation is primarily \textbf{centered} on logistics, scheduling, or administrative details of the current meeting rather than \textbf{business struggles}. \\
\hline
\end{tabular}
\caption{Comparison of original model-generated criteria and human modifications for Pain Points classification. Bold text highlights key differences from the original version.}
\label{tab:human_modifications}
\end{table*}

\section{Human-in-the-Loop Modification Example}
\label{appsec:human_modification_example}

To illustrate the nature and impact of human modifications to our generated criteria (see Section~\ref{subsec:hitl}), we present a detailed comparison of the original model-generated criteria of Sonnet 3.7 for the Pain Points classification task and the modifications made by two annotators with contrasting performance outcomes.

Table~\ref{tab:human_modifications} shows selected criteria where both annotators made modifications, comparing the original Criteria-Ex output with the human-revised versions. We focus on criteria where substantive changes were made by both annotators to highlight the different modification strategies employed. The original model achieved 73.2\% F1 on this task.

The table reveals two distinct modification approaches with contrasting outcomes. Annotator 1's modifications achieved 83.0\% F1 (+9.8\% improvement) through strategic simplification (removing verbose phrases like ``directly impacting their ability to make purchasing decisions''), semantic generalization (broadening ``purchasing process'' to ``solutions''), and scope expansion (removing overly restrictive qualifiers). These changes enhanced the criteria's applicability and clarity.

Annotator 2's modifications resulted in 71.9\% F1 (-1.3\% degradation) and involved excessive paraphrasing without semantic improvement (``prospect'' to ``potential customer'', ``challenges'' to ``struggles''), increased linguistic complexity (``directly affecting their capability'' vs ``directly impacting their ability''), and inconsistent terminology introduction. 

These findings demonstrate that effective modifications require strategic simplification, semantic broadening, and principled scope adjustments, while ineffective changes typically involve superficial rewording and unnecessary complexity. Importantly, our approach provides users with interpretable artifacts that can be meaningfully refined to improve classification performance through targeted human expertise.

\section{Average Processing Time}
\label{appsubsec:average_processing_time}

\begin{figure}[!ht]
  \includegraphics[width=\columnwidth]{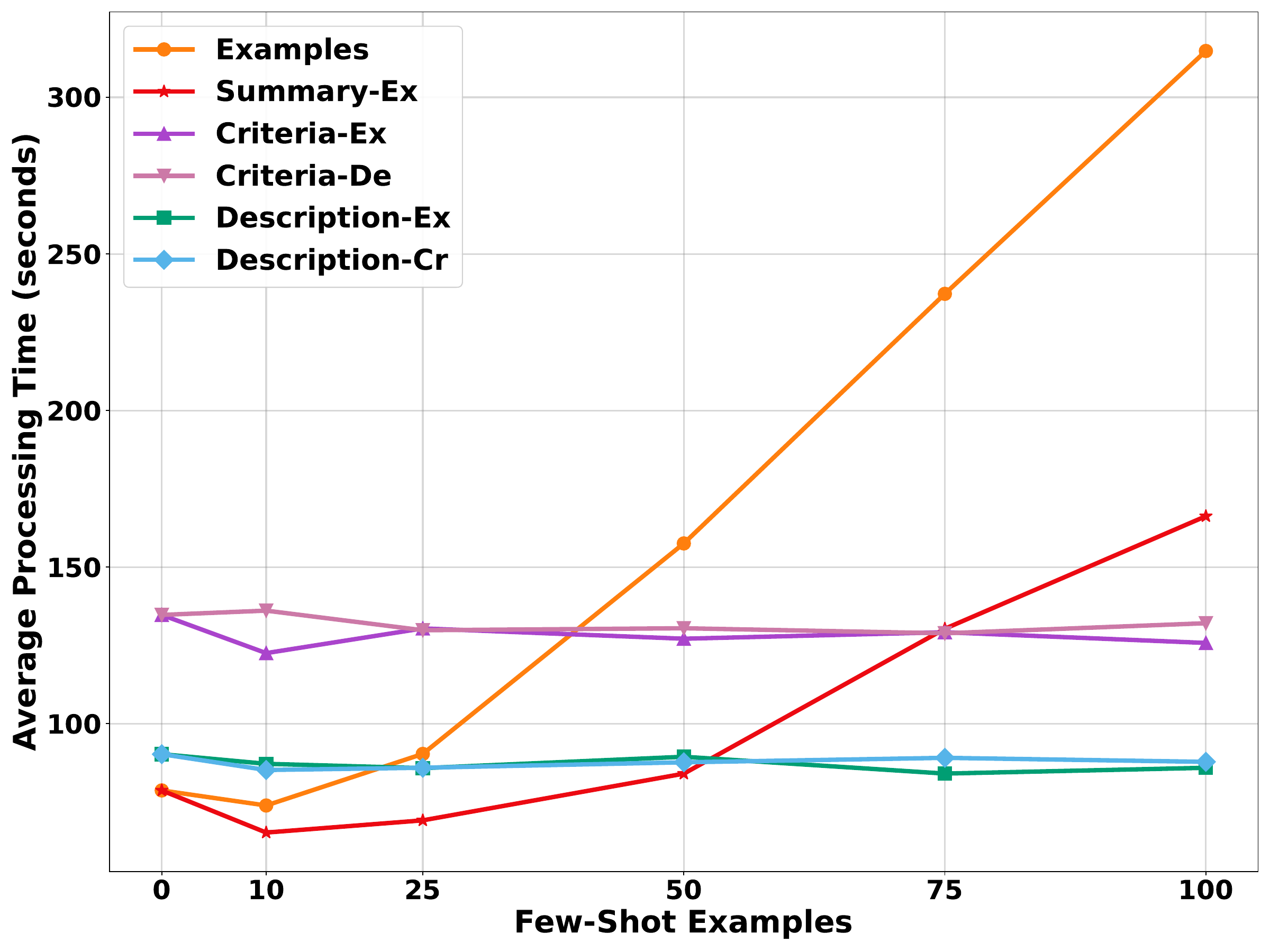}
  \caption{Average processing test time (in seconds) by method across different numbers of few-shot examples.}
  \label{fig:overall_processing_time}
\end{figure}

Figure~\ref{fig:overall_processing_time} presents the average processing time of GPT-4o across all test sets, as referenced in Section~\ref{subsec:computation_efficiency}. The processing times reported include the overhead of the knowledge extraction step (e.g., criteria, descriptions). It is important to note that for any given set of few-shot examples, our methods require only a single LLM call during the knowledge extraction phase, making this overhead minimal and required only once.

This visualization supports our findings regarding the computational efficiency advantages of our proposed methods compared to the traditional few-shot approach, especially as the number of examples increases. While Summary-Ex offers moderate improvements over the standard Examples method, the criteria and description methods scale more efficiently with increasing example counts, maintaining stable processing times regardless of the number of examples used.

\section{Comprehensive Experimental Results}
\label{appsec:full_grid}

\subsection{Main Results Grid}
\label{appsubsec:complete_grid}

\begin{figure*}[t]
  \includegraphics[width=\textwidth]{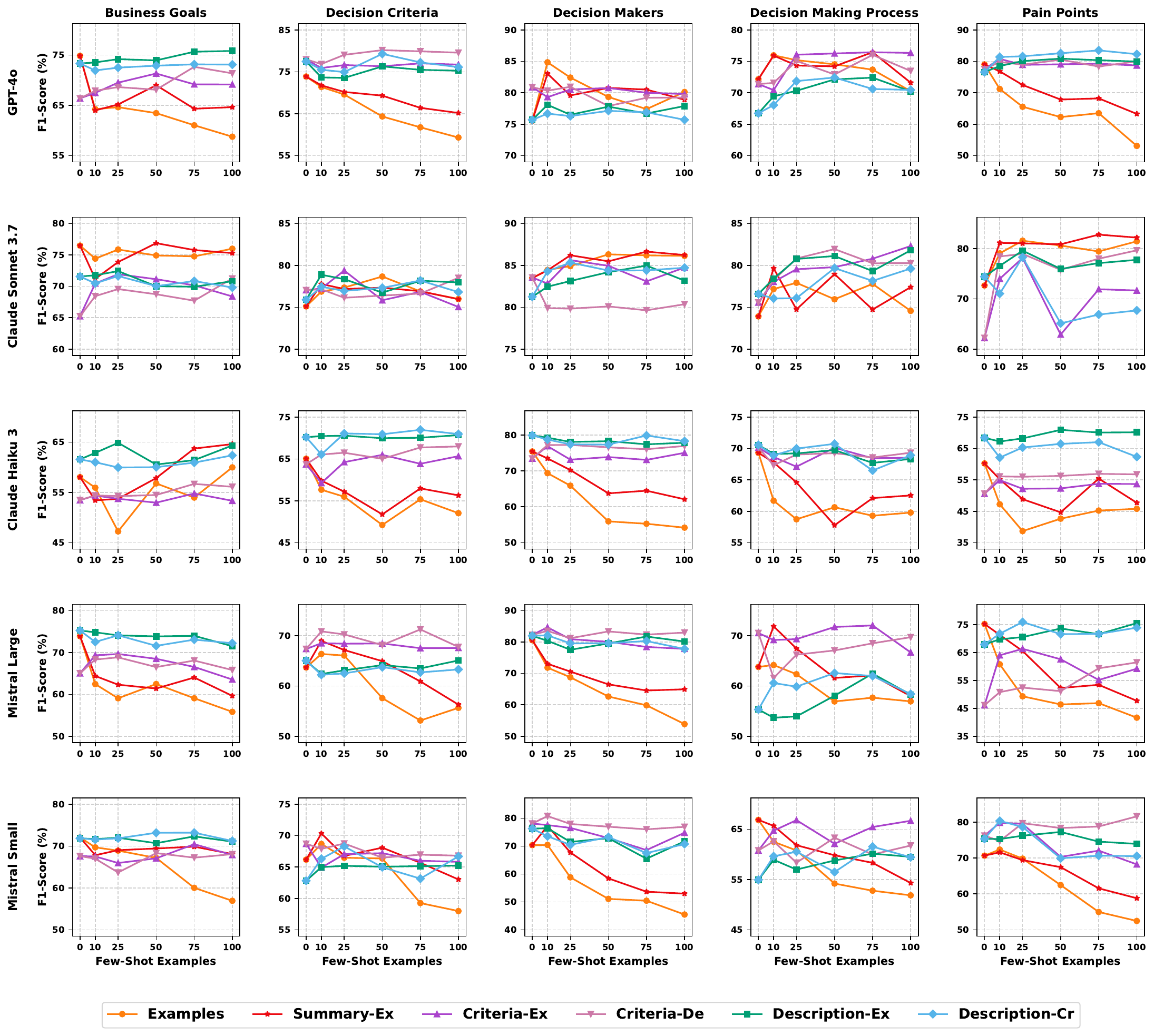}
  \caption{Macro-average F1 performance for all concepts and models.}
  \label{fig:all_experiments}
\end{figure*}

Figure~\ref{fig:all_experiments} illustrates the detailed evaluations of the macro-average F1 score, as discussed in Section~\ref{subsec:main_results}. This visualization presents performance across the five B2B concepts (Business Goals, Decision Criteria, Decision Makers, Decision Making Process, and Pain Points) and five models (GPT-4o, Claude Sonnet 3.7, Claude Haiku 3, Mistral Large, and Mistral Small). The grid layout allows for direct comparison of how each few-shot learning method performs as the number of examples increases from 0 to 100. Notably, the visualization confirms that the criteria and description methods generally maintain higher F1 scores than the traditional Examples approach and its summarized variant (Summary-Ex), both of which often exhibit declining performance with additional examples, a trend evident in all models except Claude Sonnet 3.7.

\subsection{Token-Level Compression Comparison Grid}
\label{appsubsec:compression_details}

\begin{figure*}[t]
  \centering
  \includegraphics[width=\textwidth]{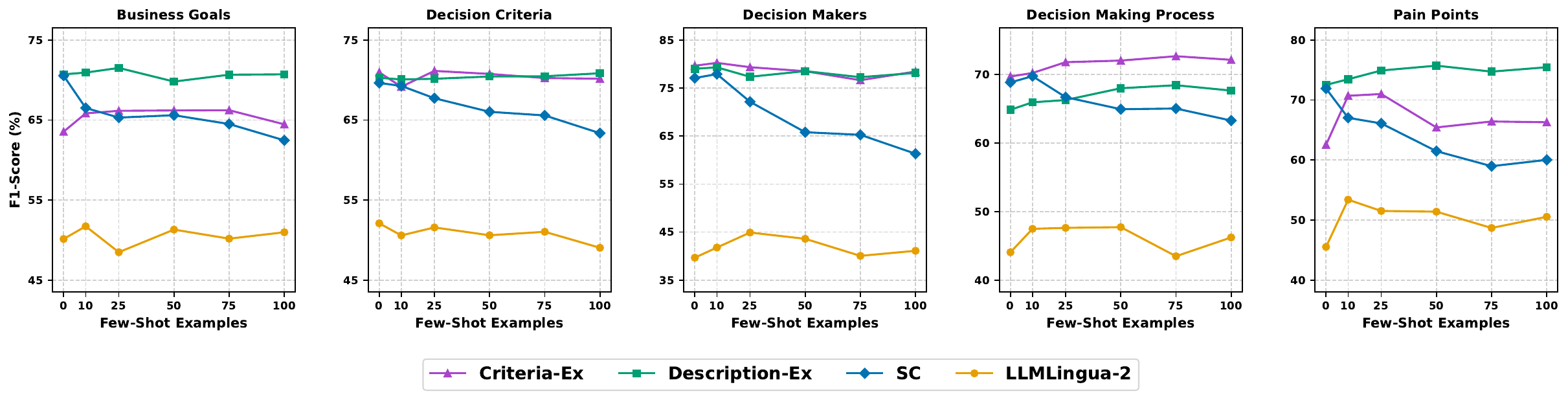}
  \caption{Macro-average F1 performance comparison with token-level compression methods, averaged over all models.}
  \label{fig:compression_baselines}
\end{figure*}

Figure~\ref{fig:compression_baselines} presents the detailed per-concept comparison with token-level compression methods, averaged across all five models. The results complement Table~\ref{tab:token_compression_comparison} in Section~\ref{subsec:token_compression_comparison}, showing consistent patterns across the five B2B concepts. Both Criteria-Ex and Description-Ex demonstrate robust performance as examples increase, while LLMLingua-2 yields poor results, consistently staying under 50\% F1 across all concepts as it compresses both task instructions and examples. SC shows competitive initial performance but exhibits consistent degradation, with steep declines observed in all concepts.

\end{document}